\documentclass[11pt,a4paper]{scrreprt} 

\usepackage[utf8]{inputenc}
\usepackage[T1]{fontenc}
\usepackage{url}
\usepackage{booktabs}
\usepackage{amsfonts}
\usepackage{nicefrac}
\usepackage{microtype}
\usepackage{xcolor}
\usepackage{mathtools}
\usepackage{amsmath}
\usepackage{amsthm}
\usepackage{tcolorbox}
\usepackage{placeins}
\usepackage[shortlabels]{enumitem}

\usepackage[round]{natbib}

\usepackage[top=3cm, bottom=3cm, left=1.5cm, right=1.5cm]{geometry}
\usepackage{color}
\usepackage{amsmath}
\usepackage{amssymb}
\usepackage{amsfonts}
\usepackage{tcolorbox}
\usepackage[english]{babel}
\usepackage{fancyhdr}
\usepackage[mathscr]{euscript}

\usepackage{authblk,amssymb,latexsym,amsfonts,amsmath}
\usepackage{color}
\usepackage{algorithm}
\usepackage{algpseudocode}
\usepackage{bm}
\usepackage{mathtools}
\usepackage[title]{appendix}
\usepackage[T1]{fontenc}
\usepackage{dsfont}
\usepackage{natbib}
\usepackage{url}
\usepackage{float}
\usepackage{placeins}

\usepackage[font=small,labelfont=bf]{caption}
\usepackage{authblk} 

\fancypagestyle{firststyle}
{
   \fancyhf{}
   
}

\renewenvironment{proof}[1][\proofname]{\noindent{\bfseries #1.} }{\qed \\ }

\DeclareMathAlphabet{\mathcal}{OMS}{cmsy}{m}{n}

\usepackage[utf8]{inputenc}
\usepackage[T1]{fontenc}
\usepackage{url}
\usepackage{booktabs}
\usepackage{nicefrac}
\usepackage{microtype}
\usepackage{xcolor}
\usepackage{float}
\usepackage{amssymb,amsmath}
\usepackage{bm}
\usepackage{mathtools}
\usepackage[round]{natbib}
\usepackage{todonotes}
\setuptodonotes{inline}
\usepackage{verbatim}
\usepackage{placeins}
\usepackage{xspace}
\setlist[itemize]{leftmargin=2em}
\allowdisplaybreaks
\DeclareMathAlphabet\mathbfcal{OMS}{cmsy}{b}{n}

\newcommand{\coremmd}{h^{\operatorname{MMD}}_k}

\newcommand{\corehsick}{h^{\operatorname{HSIC}}_{\hsickx\!,\hsicky}}

\newcommand{\hsickx}{{k^{\mathcal{X}}}}
\newcommand{\hsicky}{{k^{\mathcal{Y}}}}
\newcommand{\kxl}{{k^{\mathcal{X}}_\lambda}}
\newcommand{\kyl}{{k^{\mathcal{Y}}_\mu}}

\newcommand{\one}{\bm{1}}

\newcommand{\Xbb}{{\mathbb{X}}}
\newcommand{\Ybb}{{\mathbb{Y}}}
\newcommand{\Dcal}{{\mathcal{D}}}

\newcommand{\Ycal}{{\mathcal{Y}}}
\newcommand{\Hcal}{{\mathcal{H}}}

\newcommand{\Kcal}{{\mathcal{K}}}

\newcommand{\Lcal}{{\mathcal{L}}}
\newcommand{\Pcal}{{\mathcal{P}}}
\newcommand{\Scal}{{\mathcal{S}}}
\newcommand{\kdisc}{{\mathrm{Kdisc}}}

\newcommand{\sbm}{{\bm{s}}}
\newcommand{\sbmp}{{\bm{s}_P}}

\newcommand{\dbm}{{\bm{\delta}}}

\newcommand{\psib}{{\bm{\psi}}}
\newcommand{\E}{\mathbb{E}}
\newcommand{\R}{\mathbb{R}}

\newcommand{\disc}{{\mathrm{Disc}}}
\newcommand{\Xm}{\mathbb{X}_m}
\newcommand{\Xn}{\mathbb{X}_n}
\newcommand{\Yn}{\mathbb{Y}_n}

\def \P {\mathbb{P}}

\newcommand{\pp}[1]{\!\left(#1\right)}

\newcommand{\bigO}[1]{\mathcal{O}\!\left(#1\right)}

\newcommand{\ie}{\emph{i.e.}\xspace}
\newcommand{\eg}{\emph{e.g.}\xspace}

\newcommand{\mP}{\mathbb{P}}

\newcommand{\dd}{\mathrm{d}}

\newcommand{\LL}{\lambda_1\cdots\lambda_d}

\usepackage{mathtools}
\usepackage{xcolor}
\usepackage{lastpage}
\usepackage{enumitem}
\usepackage{bbm}
\usepackage{bm}
\usepackage{float}
\usepackage{caption}
\usepackage{setspace}
\usepackage{placeins}
\usepackage{afterpage}
\usepackage{multirow}
\usepackage{array}
\usepackage{tabularx}
\usepackage{hhline}
\usepackage{colortbl} 
\usepackage{booktabs}

\newcommand{\PPP}{{\mathbb P}}

\newcommand{\EE}[2]{{{\mathbb E}_{#1}\!\left[#2\right]}}

\newcommand{\VV}[2]{{\mathrm{var}_{#1}\!\left(#2\right)}}

\newcommand{\h}{h_\lambda}
\newcommand{\kk}{k_\lambda}

\newcommand{\p}[1]{\!\left( #1 \right)}

\DeclareMathAlphabet{\mymathbb}{U}{BOONDOX-ds}{m}{n}

\DeclarePairedDelimiter\abs{\lvert}{\rvert}
\DeclarePairedDelimiter\norm{\lVert}{\rVert}
\makeatletter
\let\oldabs\abs
\def\abs{\@ifstar{\oldabs}{\oldabs*}}
\let\oldnorm\norm
\def\norm{\@ifstar{\oldnorm}{\oldnorm*}}
\makeatother

\makeatletter
\newcommand*\bigcdot{\mathpalette\bigcdot@{.5}}
\newcommand*\bigcdot@[2]{\mathbin{\vcenter{\hbox{\scalebox{#2}{$\m@th#1\bullet$}}}}}
\makeatother

\makeatletter
\newcommand\footnoteref[1]{\protected@xdef\@thefnmark{\ref{#1}}\@footnotemark}
\makeatother

\algblockdefx{MRepeat}{EndRepeat}{\textbf{repeat}}{}
\algnotext{EndRepeat}

\makeatletter
\newcommand\fs@nobottomruled{\def\@fs@cfont{\bfseries}\let\@fs@capt\floatc@ruled
  \def\@fs@pre{\hrule height.8pt depth0pt \kern2pt}\def\@fs@post{}\def\@fs@mid{\kern2pt\hrule\kern2pt}\let\@fs@iftopcapt\iftrue}
\makeatother

\floatstyle{nobottomruled}
\restylefloat{algorithm}

\makeatletter
\AtBeginDocument{\def\HH@loop{\ifx\@tempb`\def\next##1{\the\toks@\cr}\else\let\next\HH@let
  \ifx\@tempb|\if@tempswa
          \ifx\CT@drsc@\relax
           \HH@add{\hskip\doublerulesep}\else
           \HH@add{{\CT@drsc@\vrule\@width\doublerulesep}}\fi
          \fi\@tempswatrue
          \HH@add{{\CT@arc@\vline}}\else
  \ifx\@tempb:\if@tempswa
          \ifx\CT@drsc@\relax
           \HH@add{\hskip\doublerulesep}\else
           \HH@add{{\CT@drsc@\vrule\@width\doublerulesep}}\fi
              \fi\@tempswatrue
      \HH@add{\@tempc\HH@box\arrayrulewidth\arrayrulewidth\@tempc}\else
  \ifx\@tempb;\if@tempswa
          \ifx\CT@drsc@\relax
           \HH@add{\hskip\doublerulesep}\else
           \HH@add{{\CT@drsc@\vrule\@width\doublerulesep}}\fi
              \fi\@tempswatrue
      \HH@add{\@tempc\HH@box\z@\arrayrulewidth\@tempc}\else
  \ifx\@tempb##\if@tempswa\HH@add{\hskip\doublerulesep}\fi\@tempswatrue
         \HH@add{{\CT@arc@\vline\copy\@ne\@tempc\vline}}\else
  \ifx\@tempb~\@tempswafalse
           \if@firstamp\@firstampfalse\else\HH@add{&\omit}\fi
              \ifx\CT@drsc@\relax
                \HH@add{\hfil}\else
                 \HH@add{{\CT@drsc@\leaders\hrule\@height\HH@height\hfil}}\fi
                 \else
  \ifx\@tempb-\@tempswafalse
           \gdef\HH@height{\arrayrulewidth}\if@firstamp\@firstampfalse\else\HH@add{&\omit}\fi
              \HH@add{{\CT@arc@\leaders\hrule\@height\arrayrulewidth\hfil}}\else
  \ifx\@tempb=\@tempswafalse
       \gdef\HH@height{\dimen\thr@@}\if@firstamp\@firstampfalse\else\HH@add{&\omit}\fi
       \HH@add
          {\rlap{\copy\@ne}\leaders\copy\@ne\hfil\llap{\copy\@ne}}\else
  \ifx\@tempb t\HH@add{\def\HH@height{\dimen\thr@@}\HH@box\doublerulesep\z@}\@tempswafalse\else
  \ifx\@tempb b\HH@add{\def\HH@height{\dimen\thr@@}\HH@box\z@\doublerulesep}\@tempswafalse\else
  \ifx\@tempb>\def\next##1##2{\HH@add{{\baselineskip\p@\relax
       ##2\global\setbox\@ne\HH@box\doublerulesep\doublerulesep}}\HH@let!}\else
  \PackageWarning{hhline}{\meaning\@tempb\space ignored in \noexpand\hhline argument\MessageBreak}\fi\fi\fi\fi\fi\fi\fi\fi\fi\fi\fi
  \next}
}

\usepackage[utf8]{inputenc}
\usepackage[T1]{fontenc}
\usepackage{url}
\usepackage{booktabs}
\usepackage{amsfonts}
\usepackage{nicefrac}
\usepackage{microtype}
\usepackage{xcolor}
\usepackage{subcaption}
\usepackage{microtype}
\usepackage{graphicx}
\usepackage{booktabs} 
\usepackage{amsmath}
\usepackage{amssymb}
\usepackage{bm}
\usepackage{mathtools}
\usepackage{xspace}
\usepackage{bbm}

\newcommand{\kl}{k_{\lambda}}

\usepackage[utf8]{inputenc}
\usepackage[T1]{fontenc}
\usepackage{url}
\usepackage{booktabs}
\usepackage{amsfonts}
\usepackage{nicefrac}
\usepackage{microtype}
\usepackage{xcolor}
\usepackage{mathtools}
\usepackage{amsmath}
\usepackage{tcolorbox}
\usepackage{placeins}
\usepackage{amssymb}
\usepackage{bm}
\usepackage{float}

\newcommand{\Zn}{\mathbb{Z}_N}
\newcommand{\D}{\mathcal{D}}

\newcommand{\PP}{\mathbb P}

\usepackage[utf8]{inputenc} 
\usepackage[T1]{fontenc}    
\usepackage{url}            
\usepackage{booktabs}       
\usepackage{amsfonts}       
\usepackage{nicefrac}       
\usepackage{microtype}      
\usepackage{xcolor}         

\usepackage{amssymb,amsmath}
\usepackage{bm}
\usepackage{mathtools}
\usepackage{verbatim}
\usepackage{enumitem}
\usepackage{placeins}

\usepackage{wrapfig}

\usepackage{thm-restate}

\newcommand{\etc}{\emph{etc}}

\RequirePackage[
  pdfstartview=FitH,
  breaklinks=true,
  bookmarks=true,
  colorlinks=true,
  linkcolor= blue,
  anchorcolor=blue,
  citecolor=blue,
  filecolor=blue,
  menucolor=blue,
  urlcolor=blue
  ]{hyperref}
  \AtBeginDocument{
  \hypersetup{
    pdfauthor = {Antonin Schrab},
    colorlinks = true,
    urlcolor = cyan,
    linkcolor = blue,
    citecolor = teal,
    pdftitle = {Title - compilation : \today}
  }
}
\usepackage{hyperref} 
\usepackage{cleveref}
\crefname{assumption}{Assumption}{Assumptions}
\crefname{equation}{Equation}{Equations}
\crefname{figure}{Figure}{Figures}
\crefname{table}{Table}{Tables}
\crefname{section}{Section}{Sections}
\crefname{theorem}{Theorem}{Theorems}
\crefname{lemma}{Lemma}{Lemmas}
\crefname{corollary}{Corollary}{Corollaries}
\crefname{example}{Example}{Examples}
\crefname{appendix}{Appendix}{Appendices}
\crefname{remark}{Remark}{Remarks}

\usepackage[nottoc]{tocbibind}

\newcommand{\mysection}[1]{\section{#1}}
\newcommand{\mysubsection}[1]{\subsection{#1}}

\makeatletter 
 
\@addtoreset{algorithm}{chapter} 
\makeatother

\creflabelformat{equation}{#2\textup{#1}#3}
\crefformat{footnote}{#2\footnotemark[#1]#3}

\usepackage{thmtools}
\declaretheorem[name=Theorem,numberwithin=chapter]{theorem}
\declaretheorem[name=Proposition,sibling=theorem]{proposition}
\declaretheorem[name=Corollary,sibling=theorem]{corollary}

\declaretheorem[name=Lemma,sibling=theorem]{lemma}

\counterwithout*{footnote}{chapter}

\usepackage{appendix}
\renewcommand\thesection{\arabic{section}}
\counterwithout{equation}{chapter} 

\begin{document}

\thispagestyle{firststyle}
{\center

	{
		\fontsize{32}{52}\selectfont 
		\vspace{2cm}
		\textbf{A Unified View of}\\[-0.5cm]
		\textbf{Optimal Kernel Hypothesis Testing }\\[1cm]
	}

	{
		\fontsize{16}{14}\selectfont 
		\textbf{Antonin Schrab} \\[0.3cm]
		\textit{Centre for Artificial Intelligence } \\[0.2cm]
		\textit{Gatsby Computational Neuroscience Unit} \\[0.2cm]
		\textit{University College London} \\[0.3cm]
	}

}

\vskip 0.5cm
{\LARGE\centerline{\textbf{Abstract}}}
\vskip 0.3cm

\begin{center}
\begin{minipage}{10.0cm}
This paper provides a unifying view of optimal kernel hypothesis testing across the MMD two-sample, HSIC independence, and KSD goodness-of-fit frameworks. Minimax optimal separation rates in the kernel and $L^2$ metrics are presented, with two adaptive kernel selection methods (kernel pooling and aggregation), and under various testing constraints: computational efficiency, differential privacy, and robustness to data corruption. Intuition behind the derivation of the power results is provided in a unified way across the three frameworks, and open problems are highlighted.
\end{minipage}
\end{center}
\bigskip
\bigskip

This paper corresponds to the main chapter of my PhD thesis \citep[Chapter 3]{schrab2025optimal}, it provides a unifying view of optimality testing results using the kernel discrepancies: Maximum Mean Discrepancy (MMD), Hilbert--Schmidt Independence Criterion (HSIC), and Kernel Stein Discrepancy (KSD). 
The focus of this paper is on hypothesis testing, we refer the reader to \citet{schrab2025practical} for a detailed introduction to these kernel discrepancies, to their estimators, and to kernel pooling---see \citealp{schrab2025optimal} for the complete thesis containing both the kernel discrepancy introduction and the unified optimality results in a single document.

Statistical hypothesis testing plays a crucial role in machine learning, and more generally in all sciences, as it allows to rigorously guarantee that some patterns observed in the data are statistically significant (\ie, not simply due to chance).
While many tests have been designed to test for specific distributions, we focus on the more general setting of non-parametric testing which imposes no distributional assumption on the data.
Kernel methods have become a well-established powerful toolbox to tackle non-parametric testing problems such as the two-sample, independence, and goodness-of-fit problems.

In \Cref{sec:def}, we formalise the hypothesis testing framework, highlighting test level and power properties.
In \Cref{sec:multiple_testing}, we present multiple testing via the aggregation method, which is strictly more powerful than using a Bonferroni correction.
In \Cref{sec:dp_dc}, we consider hypothesis testing under three constraints: computational efficiency, differential privacy and robustness to data corruption.
In \Cref{sec:two_sample_testing,sec:independence_testing,sec:gof_testing}, we introduce the two-sample, independence and goodness-of-fit testing frameworks, respectively, and derive power guarantees under kernel and $L^2$ uniform separation in the standard, efficient, private and robust settings, with aggregation and pooling kernel adaptation methods.
In \Cref{sec:future_work}, we highlight open problems which are left for future work.
In \Cref{sec:proof_sketches}, we provide intuition behind the proofs of all the power results presented.

Thoughout this chapter, we use the notation $a\lesssim b$ when there exists a constant $C>0$ such that $a\leq Cb$ (similarly for $\gtrsim$), and write $a\asymp b$ if $a\lesssim b$ and $a\gtrsim b$.

\newpage
\thispagestyle{empty}

\begin{figure}[H]
\vspace{-1.9cm}
	\center\includegraphics[width=\textwidth]{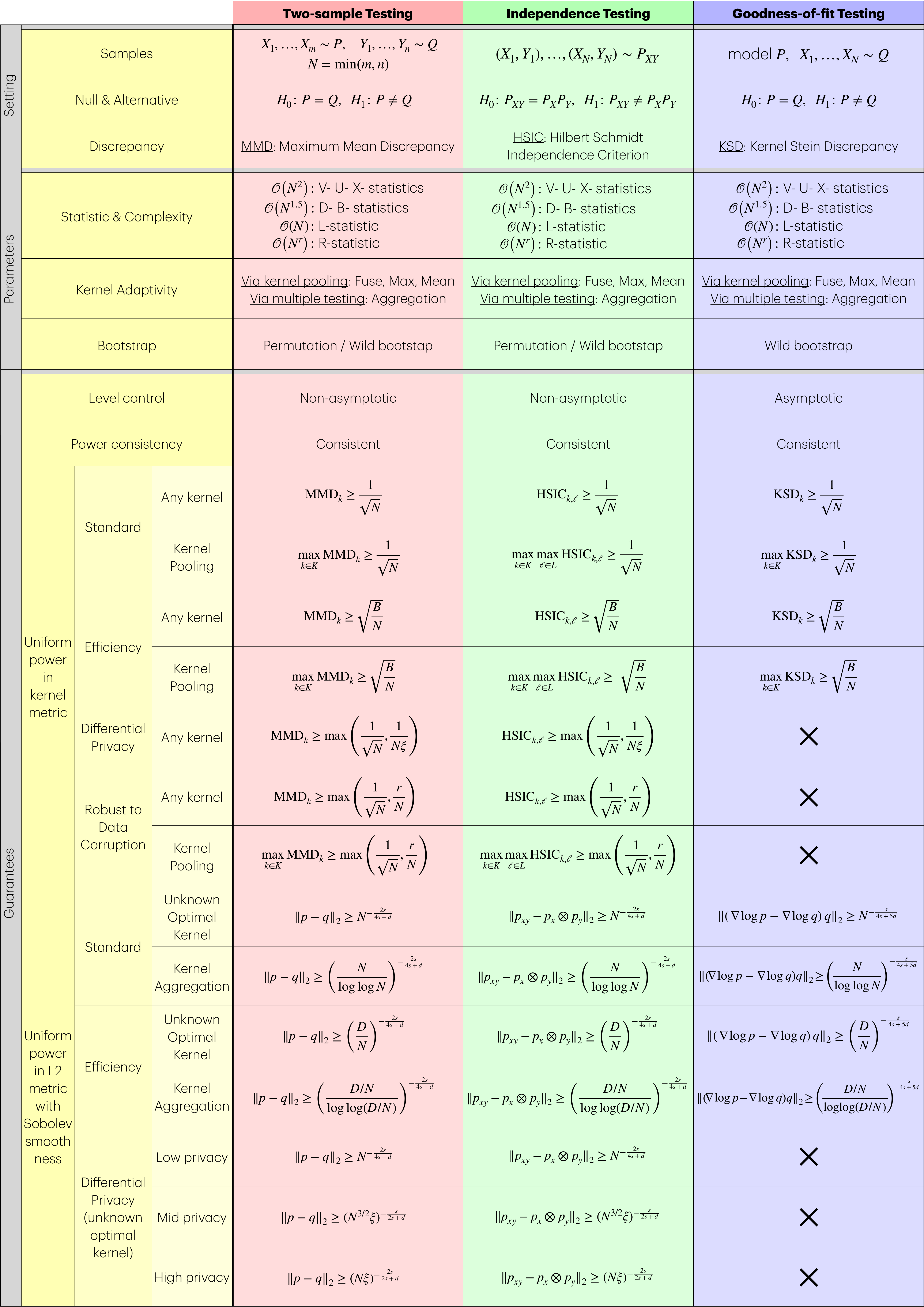}
	\caption{
	\label{fig:main}
	\!\!Uniform separation rates in the sample size $N$. 
	\emph{Efficiency:} block size $B$ ranging from $1$ to $N$, and design size $D$ typically ranging from $N$ to $N^2$. 
	\emph{Differential privacy:} $(\varepsilon,\delta)$-DP with $\xi = \varepsilon + \log\!\big({1}/({1-\delta})\big)$.
	\emph{Robust to data corruption:} robust to the corruption of up to $r$ samples.
	\emph{Sobolev regularity:} smoothness $s$ and dimension $d$.
	All uniform separation rates presented hold with logarithmic dependencies in the test errors.
	}
\end{figure}

\mysection{Hypothesis testing}
\label{sec:def}

We formally introduce the concept of hypothesis testing in \Cref{subsec:def_hp}, we discuss test level control in \Cref{subsec:level_hp} and test power guarantees in \Cref{subsec:power_hp}.

\subsection{Definition of hypothesis testing}
\label{subsec:def_hp}

\paragraph{Statistical hypothesis testing.}
Consider a space of distributions $\mathcal{P}$ partitioned into disjoint subsets $\mathcal{P}_0$ and $\mathcal{P}_1$.
Given some samples drawn i.i.d.~from $P \in \mathcal{P}$, the aim of hypothesis testing is to test whether the null $\Hcal_0$ or the alternative $\Hcal_1$ holds, where
\begin{equation}
\mathcal{H}_0\colon P \in \mathcal{P}_0
\qquad 
\qquad
\textrm{ and } 
\qquad
\qquad
\mathcal{H}_1\colon P \in \mathcal{P}_1.
\end{equation}
Typically, the null set $\Pcal_0$ is much smaller than $\Pcal_1$, and consists of distributions satisfying a specific property, which we aim to obtain statistical evidence against.

\paragraph{Examples.} 
For two-sample testing, the space $\Pcal$ consists of all pairs of distributions, and $\Pcal_0$ of only the pairs with the same distribution for each component.
For independence testing, the space $\Pcal$ consists  of joint distributions, and $\Pcal_0$ of the ones which are equal to the product of their marginals.
For goodness-of-fit testing, we have a reference model distribution $P_{\textrm{model}}$, the space $\Pcal$ consists of all distributions while $\Pcal_0 = \{P_{\textrm{model}}\}$ is simply the model distribution.
For more details on these testing frameworks, see \Cref{sec:two_sample_testing,sec:independence_testing,sec:gof_testing}.

\paragraph{Hypothesis test.}
A test is a function which takes as input the i.i.d.~samples from $P\in\Pcal$, and returns 0 if the null $\Hcal_0$ is believed to hold, or 1 otherwise (\ie $\Hcal_1$ is believed to hold).
A test is usually constructed by first computing a statistic, a real value computed from the data which is designed to capture evidence against the null when it exists.
This statistic is then compared to a rejection region to decide whether to reject the null.
Equivalently, the test can be defined via its p-value rather than via its quantile (see details below in \Cref{subsec:level_hp}).

\paragraph{Hypothesis tests using discrepancies.}
Often, the null set $\Pcal_0$ corresponds exactly to distributions for which a well-chosen discrepancy equals zero, as illustrated in the above examples (\eg MMD for two-sample testing, HSIC for independence testing, KSD for goodness-of-fit testing).
The test statistic is then an estimator of the corresponding discrepancy, and the test is rejected when the statistic is greater than some rejection threshold.

\subsection{Level of hypothesis testing}
\label{subsec:level_hp}

\paragraph{Level: type I error.}
The rejection threshold is usually chosen such that the probability of rejecting the null when it actually holds (\ie, type I error) is at most $\alpha$ (typically $5\%$) uniformly over $\Pcal_0$.
In which case, we say that the test has level $\alpha\in(0,1)$, that is
\begin{equation}
\sup_{P_0\in\Pcal_0} \P_{P_0}(\textrm{reject } \Hcal_0) ~\leq~ \alpha.
\end{equation}
Typically, we would want this inequality to be tight: if the type I error is required to be less than $\alpha$, we would ideally want it to be as close as possible to $\alpha$ since a smaller type I error (\eg, conservative test) would result in a larger type II error (see \Cref{subsec:power_hp}).

\paragraph{Test construction (quantile and p-value).}
To construct a test which has level control at $\alpha$, we rely on two bootstrap methods\footnote{In some cases, the distribution of the test statistic under the null can be known (either directly or asymptotically), in which case, the test threshold can be set to its known $(1\!-\!\alpha)$-quantile directly.} (permutations and wild bootstrap) introduced below, which are used to simulate values of the statistic under the null (either non-asymptotically or asymptotically).
Using either of these methods, we can compute many bootstrapped statistics. 
The test can then be constructed either via the quantile point of view, or via the p-value point of view, which are equivalent \citep[Lemma 17]{kim2023differentially}.
For the quantile case, the test threshold is set to be the $(1-\alpha)$-quantile of all the statistics (original and bootstrapped) which simulate the null, the test is then defined as rejecting the null if the test statistic is strictly larger than the test threshold.
Equivalently, the p-value can be computed as the proportion of statistics (original and bootstrapped) which are smaller of equal to the original statistic (see \citealp[Equation 3]{kim2023differentially} for a formal definition), the test is then defined as rejecting the null if the p-value is smaller or equal to $\alpha$.
These procedures result in a test with type I error exactly equal to $\alpha$ under mild conditions \citep[Lemma 15]{kim2023differentially}.

\paragraph{Exchangeability \& permutations.}
A sample is said to be exchangeable if, for any permutation, the joint distribution over the permuted samples is equal to the joint distribution over the original samples.
Some hypothesis testing problems can be framed as testing whether exchangeability holds \citep[][Chapter 15.2]{lehmann2005testing}.
Hence, the null hypothesis can be simulated by permuting samples, and the permutation test can then be constructed by following the above construction with randomly sampled permutations.
The resulting test can then be guaranteed to control the type I error non-asymptotically at the desired level $\alpha$ (\citealp[Lemma~1]{romano2005exact}, see also \citealp[Lemma 15]{kim2023differentially}).
As explained in \Cref{subsec:twosample_main,subsec:ind_main,subsec:gof_main}, the two-sample and independence problems can be framed as testing for exchangeability, but the goodness-of-fit one cannot.

\paragraph{Wild bootstrap.}
While the permutation approach presented above is applicable when using any type of statistic. 
We now present the wild bootstrap method \citep{wu1986jackknife} which is specifically designed for one-sample second-order statistics\footnote{The wild bootstrap can also be defined more generally for higher order (and even two-sample) $V$-statistics and its incomplete counterparts \citep[Equations 2 and 4]{chwialkowski2014wild}, however, we use it only with one-sample second-order statistics in this work.}
$
	{|\Dcal|}^{-1}\sum_{(i,j) \in \Dcal} h(X_i, X_j)
$
for some design $\Dcal\subseteq \{(i,j):1\leq i,j\leq n\}$ and some core function $h$, for data $X_1,\dots,X_n$.
As discussed in \citet[Section 3]{schrab2025practical}, all three kernel discrepancies (\ie, MMD, HSIC, KSD) admit various estimators of this form, each with different computational complexities.
A wild bootstrapped statistic is then expressed as
\begin{equation}
\label{eq:wild_bootstrap_statistic}
\frac{1}{|\Dcal|}\sum_{(i,j) \in \Dcal} \varepsilon_i \varepsilon_j h(X_i, X_j)
\end{equation}
where $\varepsilon_1,\dots,\varepsilon_n$ are i.i.d.~Rademacher variables\footnote{We focus on this setting, while the wild bootstrap can more generally be used with any distribution having mean zero and variance one (\eg, standard Gaussian).}, \ie, each taking value $+1$ or $-1$ with probability $\nicefrac{1}{2}$.
Under the null hypothesis, with some assumptions on the core function $h$, the asymptotic distribution of the wild bootstrapped statistic can be proven to be the same as the one of the original statistic \citep[Theorem 1]{chwialkowski2014wild}.
So, the wild bootstrap can be used to simulate the null distribution asymptotically.
Hence, following the test construction depicted above leads to a test which controls the type I error at level $\alpha$ asymptotically.
For some frameworks testing for exchangeability and with particular choices of the core function, computing the wild bootstrapped statistic of \Cref{eq:wild_bootstrap_statistic} can be equivalent to computing the statistic on permuted samples for some specific permutation (see \citealp[Appendix B]{schrab2021mmd} for the MMD two-sample case, and \citealp[Appendix F.1]{schrab2022efficient} for the HSIC independence case). 
Leveraging these results, we obtain non-asymptotic type I error control for these wild bootstrap hypothesis tests.

\paragraph{Computational complexity: permutations and wild bootstrap.}
The computational complexities of the tests based on either permutations or wild bootstrap are of the order of the cost of computing the statistic times the number of bootstrapped statistics.
It is common to use a large number of bootstrapped statistics (\eg, 500, 1000, 2000) but this is not necessary: our results (\eg, \citealp[Theorem 7]{kim2023differentially}) require only 192 of them (for $\alpha=\beta=0.05$), and \citet[Theorem 1, Section 5]{domingo2023compress} uses only $39$ permutations as motivated by their theory.
While computing these bootstrapped statistics many times might seem computationally expensive, it is often possible to reduce the runtimes drastically.
Firstly, the bootstrapped statistics can all be computed in parallel if needed.
Secondly, when using estimators of kernel discrepancies, the kernel/core values need to be computed only once. 
Thirdly, it is sometimes possible to vectorise the computation of all bootstrapped statistics, which then results in significant speed-ups: this is always possible for the wild bootstrap, and is possible for permutations when using the MMD but not when using the HSIC \citep[Appendix C]{schrab2021mmd}.

\paragraph{Permutations vs wild bootstrap.}
If the framework considered is not equivalent to testing exchangeability, then the permutation method does not apply, and one should rely on the wild bootstrap approach and its asymptotic level control (\eg, goodness-of-fit testing in \Cref{sec:gof_testing}).
When testing for exchangeability (\eg, two-sample and independence testing in \Cref{sec:two_sample_testing,sec:independence_testing}), a clear advantage of the permutation method is that it can be applied when using any statistic, while the wild bootstrap method only works for one-sample second-order $V$-statistics and its incomplete variants (\ie, not for the two-sample MMD V- and U-statistics \citep[\emph{e.g.},][Equations 7 and 12]{schrab2025practical} for the case $m\neq n$, and not for the full fourth-order HSIC V- and U-statistics \citep[\emph{e.g.},][Equations 21 and 26]{schrab2025practical}).
So, when using a one-sample second-order statistic, both methods can be used, and for two-sample and independence testing, using a wild bootstrap is equivalent to using a subset of permutations (\citealp[Appendix B]{schrab2021mmd} and \citealp[Appendix F.1]{schrab2022efficient}), so both methods benefit from non-asymptotic level guarantees in that setting.
When using a complete one-sample second-order statistic (\ie, U-statistics or V-statistics), then either method can be used and perform similarly.
When using an incomplete one-sample second-order statistic (see \citealp[Section 3]{schrab2025practical}), using a wild bootstrap is highly preferable for computational reasons, as computing a permuted incomplete statistic likely results in having to evaluate the core function at new pairs of data points (not included in the original design but belonging to the permuted design).

\subsection{Power of hypothesis testing}
\label{subsec:power_hp}

\paragraph{Type II error.}
Having type I error control at level $\alpha$ holding uniformly across $\Pcal_0$, we would ideally also like to control the type II error (\ie, failing to reject the null when the alternative holds) by $\beta$ uniformly over $\Pcal_1$.
However, this is known to be impossible as both types of errors cannot be minimised simultaneously. 

\paragraph{Pointwise power.}
While type II error control cannot hold uniformly over $\Pcal_1$, it is possible to guarantee pointwise power (also known as pointwise consitency):
for some fixed $P_1\in\Pcal_1$, the power (\ie, 1 minus the type II error) converges to 1, that is
\begin{equation}
\lim_{n\to\infty} \P_{P_1}(\textrm{reject } \Hcal_0)=1.
\end{equation}

\paragraph{Uniform power.}
Given a level $\alpha$ test, it is also possible to guarantee high power uniformly over some strictly smaller subset $\Scal_1$ of alternatives from $\Pcal_1$ in the sense that, for some $\beta\in(0,1)$, we have
\begin{equation}
\label{eq:uniform_power}
\sup_{P_1\in\Scal_1} \P_{P_1}(\textrm{reject } \Hcal_0) ~\geq~ 1 - \beta.
\end{equation}
We now present a class of subsets for $\Scal_1$ which are separated from the null distributions.
Let $\disc$ be a discrepancy which is zero exactly for the null distributions of $\Pcal_0$ (it does not necessarily need to be the same discrepancy as the one estimated for the test statistic).
Then, a candidate for the subset for which uniform power holds takes the form of all distributions satisfying 
$
\disc \geq \rho
$
for some positive separation rate $\rho$, possibly with some additional regularity condition on the distributions.
For a fixed discrepancy $\disc$, the aim is then:
\begin{enumerate}
\item \textbf{Upper bound:} For a given level $\alpha$ test, to determine the smallest separation rate $\rho$ for which uniform power holds (\Cref{eq:uniform_power}).
\item \textbf{Lower bound:} To determine the largest separation rate $\rho$ such that no level $\alpha$ test can achieve uniform power as in \Cref{eq:uniform_power}.
\end{enumerate}
The uniform separate rate $\rho$ is expressed in terms of the errors $\alpha$, $\beta$, of the sample size $n$, possibly of the dimension $d$, and of any other regularity parameters introduced.
If the rates for the upper and lower bounds match, we say that a test achieving this rate is \textbf{minimax optimal} with respect to that discrepancy, and we refer to it as the \textbf{minimax rate}.
Common choices for the $\Scal_1$ discrepancies are the associated kernel discrepancies (\ie, MMD, HSIC, KSD), or the $L^2$-norm of the difference in densities (two-sample), or in score (goodness-of-fit), or between the joint and product of marginals (independence).

\paragraph{Sobolev regularity.}
An example of regularity constraint is to assume smoothness of some function capturing departures from the null (\eg, difference in densities, or in scores, or between the joint and the product of the marginals).
We characterise it via Sobolev regularity with positive smoothness $s$ \citep{adams2003sobolev}, which requires the real-valued function to be integrable and square-integrable on $\R^d$ (\ie, to belong to $L^1(\R^d)\cap L^2(\R^d)$), and to satisfy\footnote{This is a simplified definition corresponding to a Sobolev ball of radius 1 (see \citealp[Equation 1]{schrab2021mmd}).}
\begin{equation}
\label{eq:sobo}
\int_{\R^d} \big\|\xi\big\|^{2s}_2 \,\big|\widehat f(\xi)\big|^2 \,\dd \xi ~\leq~ (2\pi)^d
\end{equation}
where the Fourier transform is $\widehat f(\xi) \coloneqq \displaystyle\int_{\R^d} f(x) e^{-ix\!^\top\!\xi} \,\dd x$ for all $\xi\in\R^d$. 
Intuitively, if the parameter $s$ were to be zero, then, by Plancherel's theorem, \Cref{eq:sobo} would hold as long as $\|f\|_{L^2} \leq 1$ which would not impose any smoothness constraint.
If the parameter $s$ is large, however, then the term $\|\xi\|_2^{2s}$ in \Cref{eq:sobo} ensures that the Fourier transform decays at a rapid rate,
which imposes a smoothness requirement.

\mysection{Adaptive methods: aggregation multiple testing and kernel pooling}
\label{sec:multiple_testing}

\paragraph{Multiple testing.}
For some given hypotheses $\Hcal_0$ and $\Hcal_1$, suppose we have a test $T^{(k)}_\alpha$ with level $\alpha$ parametrised by some parameter $k$ (\eg, a kernel).
Then, we can run multiple tests $T^{(k_1)}_{\tilde \alpha},\dots,T^{(k_{|\Kcal|})}_{\tilde \alpha}$ for various parameters belonging to some finite collection $\Kcal$, each with adjusted level $\tilde\alpha\in(0,1)$ to be determined.
If one of the tests rejects the null hypothesis, then we have evidence against the null, so we should indeed reject the null.
We stress that the parameter collection $\Kcal$ needs to be fixed \emph{a priori}, or in a permutation-invariant manner when using permutation tests (see \citealp[Section 3]{biggs2023mmd} for a detailed explanation).
When the parameter $k$ corresponds to a kernel, a common choice of kernel collection \citep[\emph{e.g.},][Equation 81]{schrab2025practical} consists in Gaussian and Laplace kernels, each with ten bandwidths chosen to span the set of inter-sample distances; this choice depends on the data only in a permutation-invariant manner.
In this case, multiple testing allows for the test to be adaptive to the kernel choice.

\paragraph{Bonferroni.}
Each single test controls the probability of type I error at level ${\tilde\alpha}$ (to be determined), this means that, under the null, each single test rejects with probability at most ${\tilde\alpha}$, that is
\begin{equation}
\sup_{P_0\in\Pcal_0} \P_{P_0}\Big(T^{(k)}_{{\tilde\alpha}}\textrm{ rejects } \Hcal_0 \Big) ~\leq~ {\tilde\alpha}
\end{equation}
for each $k\in\Kcal$.
When using multiple testing, we reject the null when any of the $|\Kcal|$ tests rejects, the probability of this happening under the null can be controlled by ${\tilde\alpha}|\Kcal|$ using a union bound.
This means that running all $|\Kcal|$ tests with adjusted level ${\tilde\alpha} \coloneqq \alpha/|\Kcal|$ leads to a multiple test controlling the type I error at the desired level $\alpha$.
This is called the Bonferroni correction, with level satisfying
\begin{equation}
\sup_{P_0\in\Pcal_0} \P_{P_0}\Big(T^{(k)}_{\alpha/|\Kcal|}\textrm{ rejects } \Hcal_0 \textrm{ for some } k\in\Kcal\Big) ~\leq~ \alpha.
\end{equation}

\paragraph{Aggregation.}
The Bonferroni correction is a worst-case scenario which holds even if all the $|\Kcal|$ rejection events are disjoint (the union bound is then tight).
This is clearly not the case in this setting in which the rejection events are extremely likely to overlap (\eg, same test with different kernel).
Hence, the conservative Bonferroni level correction can be improved while maintaining type I error control at level $\alpha$.
This can be done with the aggregation method \citep{romano2005exact,romano2005stepwise,schrab2021mmd} which, for the data distribution $P\in\Pcal$, estimates the largest adjusted level $\tilde\alpha_P$ between $\alpha/|\Kcal|$ and $\alpha$ such that the multiple test correctly controls the level at $\alpha$, that is
\begin{equation}
\label{eq:aggregation_correction}
{\tilde\alpha}_P \coloneqq
\sup\Big\{u\in[\alpha/K,\alpha] \colon \P_{\pi(P)}\Big(T^{(k)}_{u}\textrm{ rejects } \Hcal_0 \textrm{ for some } k\in\Kcal\Big) \leq \alpha\Big\}
\end{equation}
where $\pi(P)$ is a permuted version of $P$ lying in $\Pcal_0$ when the null corresponds to testing for exchangeability.\footnote{When the null does not correspond to testing for exchangebaility (\eg, goodness-of-fit testing), the aggregation procedure can be used with a wild bootstrap, and the type I error control guarantees hold asymptotically.}
In that case, since $\pi(P_0) = P_0$ for all $P_0\in\Pcal_0$, we have
\begin{equation}
\P_{P_0}\Big(T^{(k)}_{
{\tilde\alpha}_{P_0}
}\textrm{ rejects } \Hcal_0 \textrm{ for some } k\in\Kcal\Big) ~\leq~ \alpha ~\textrm{ for all }~ P_0\in\Pcal_0,
\end{equation}
which, under some regularity assumptions on the null space $\Pcal_0$, can imply that
\begin{equation}
\label{eq:aggregation_level_control}
\sup_{P_0\in\Pcal_0} \P_{P_0}\Big(T^{(k)}_{
{\tilde\alpha}_{P_0}
}\textrm{ rejects } \Hcal_0 \textrm{ for some } k\in\Kcal\Big) ~\leq~ \alpha.
\end{equation}
The probability $\P_{\pi(P)}$ in \Cref{eq:aggregation_correction} can be estimated with a Monte-Carlo procedure by permuting the samples (or via wild bootstrap), and the supremum in \Cref{eq:aggregation_correction} can be estimated using a bisection method.
The aggregation level control of \Cref{eq:aggregation_level_control} still holds non-asymptotically when using these estimated quantities \citep[Proposition 8]{schrab2021mmd}.
The aggregation procedure results in the most powerful multiple test which controls the type I error at level $\alpha$, this test always outperforms the multiple test with Bonferroni correction as $\tilde\alpha_P$ is always greater or equal to $\alpha$. 
Moreover, in \Cref{eq:aggregation_correction}, it is possible to attribute some different weight $w_k$ (all summing to a quantity less or equal to one) to each kernel $k\in\Kcal$ in the level correction $u$.
See implementation details in \citet[Algorithm 1 and Section 3.5]{schrab2021mmd}.
Working with the p-value view, the aggregation procedure can be linked to the method of \citet{shah2018goodness}; detailed connections are left for future work.

\paragraph{Kernel pooling.}
Another method to construct an adaptive test is to use an adaptive estimator in the first place.
We call this kernel pooling \citep[Section 4]{schrab2025practical}: given a collection of estimators $S_k$ for $k\in\Kcal$, the pooled estimator is defined as $\textrm{pool}_{k\in\Kcal}~ S_k/\sigma_k$, where $\sigma_k$ is 1 for the unnormalised case, or is defined as in \citet[Equation 73]{schrab2025practical} for the normalised case. 
The pooling function `pool' can be chosen to be the mean, the maximum, or fuse \citep[Equations 74, 76 and 77]{schrab2025practical}.
In practice, we recommend using the fuse pooling function, which is a soft maximum with a logsumpexp expression, as it empirically leads to a more powerful test. 
Fuse pooling is analysed in details in \citet{biggs2023mmd}.
The collection of kernels can be taken to be the same as the aforementioned one used for aggregation (\emph{e.g.}, \citealp[Equation 81]{schrab2025practical}).
To conclude, adaptive tests can be constructed either via aggregation or via kernel pooling.

\mysection{Testing constraints: efficiency, privacy \& robustness}
\label{sec:dp_dc}

\paragraph{Computational efficiency.}
The kernel tests with complete estimators run in quadratic time, which can be prohibitive for very large datasets.
As such, it can be interesting to construct tests with lower computational complexity.
Typical approaches to reduce runtimes include relying on Nystr\"om approximation \citep{zhang2018large,cherfaoui2022discrete} or on random Fourier features \citep{zhang2018large,zhao2015fastmmd,chwialkowski2015fast}.
\citet{domingo2023compress} propose another interesting approach to constructing an efficient MMD test by relying on kernel thinning \citep{dwivedi2021kernel} which still achieves the minimax separation in the kernel metric for alternatives with a certain decay.
We choose to focus instead on incomplete statistics, introduced in details in \citet[Section 3]{schrab2025practical}, and to study the trade-off between computational efficiency and test power via uniform separation rates.

\paragraph{Differential privacy.}
In practice, hypothesis tests are often used on sensitive data such as medical records, personally identifiable information, facial recognition, \etc. \!(\citealp{apple2017,erlingsson2014rappor,ding2017collecting}; see \citealp[Section 1]{kim2023differentially} for further relevant discussions and references).
This can cause privacy issues, to address this we design differentially private tests which guarantee user privacy.
A randomised test (\eg, using some random noise) is said to be $(\varepsilon,\delta)$-differentially private \citep{dwork2014algorithmic} if
\begin{equation}
\begin{aligned}
	\mP \bigl( \textrm{reject } \Hcal_0 \textrm{ using }\mathbb{X}_n\bigr) ~&\leq~ e^{\varepsilon} \,\mP \bigl( \textrm{reject } \Hcal_0 \textrm{ using }\widetilde{\mathbb{X}}_n\bigr) + \delta, \\
	\mP \bigl( \textrm{fail to reject } \Hcal_0 \textrm{ using }\mathbb{X}_n\bigr) ~&\leq~ e^{\varepsilon} \,\mP \bigl( \textrm{fail to reject } \Hcal_0 \textrm{ using }\widetilde{\mathbb{X}}_n\bigr) + \delta, 
\end{aligned}
\end{equation}
for any two datasets $\mathbb{X}_n$ and $\widetilde{\mathbb{X}}_n$ differing only in one entry, for $\varepsilon>0$ and $\delta\in[0,1)$, and where the probability is taken with respect to the randomness of the test (\ie, not with respect to the data). 
See \citet[Definition 1]{kim2023differentially} for a more general definition.
Intuitively, differential privacy guarantees that the probability of a given test output remains roughly the same when the data of a single user is modified, hence guaranteeing user privacy.
We propose a procedure in \citet[Algorithm 1]{kim2023differentially} to privatise any permutation test by relying on the Laplace mechanism (see \citealp[Definition 3]{kim2023differentially}) to inject noise in every permuted statistic.
A naive application of the Laplace mechanism would see the Laplacian noise scale linearly with the number of permutations and would render the test obsolete.
Instead, we prove in \citet[Theorem 2]{kim2023differentially} that differential privacy can still be guaranteed when the noise is scaled only by a small factor of 2 (independent of the number of permutations).
By deriving tight upper bounds on the global sensitivity of the MMD and HSIC V-statistics \citep[Lemmas 5 and 6]{kim2023differentially}, we can leverage our privatisation procedure to construct $(\varepsilon,\delta)$-differentially private dpMMD and dpHSIC tests \citep[Sections 4.1 and 4.2]{kim2023differentially}.

\paragraph{Robust to data corruption.}
Another practical problem is the one of corrupted data, in real-world applications it is often the case that a portion of the data does not actually follow the distributions we would like to test (\eg, fake data, adversarially corrupted data, \etc.).
In some cases, these outliers are actually important and we want the tests to able to detect them.
In other cases, these are just noise that we wish to be robust against in order to test the real problem.
The aim of robust testing is then to test the null hypothesis when at most $r$ samples have been corrupted, possibly in an adversarial manner (this setting is more general than Huber's contamination model).
That is, the new `robust null hypothesis' is that the condition $\Hcal_0$ holds for at least $N-r$ of the $N$ samples, where the robustness parameter $r$ is specified by the user in advance depending on the testing setting and the amount of robustness required.
See \citet[Section 2.1]{schrab2024robust} for details on the robust testing framework.
While we show that differentially private tests with adjusted level can be robust \citep[Algoritm 2]{schrab2024robust}, our main contribution is to propose a new procedure to robustify any permutation test \citep[Algoritm 1]{schrab2024robust} by adding a factor of $2r\Delta$ to the rejection threshold (\ie, quantile obtained using permutations), where $\Delta$ is the global sensitivity of the test statistic \citep[Definition 2]{kim2023differentially}.
Using these procedures, 
robust two-sample and independence tests, dcMMD and dcHSIC, can be constructed, which are robust up to $r$ corruption.

\mysection{Two-sample testing}
\label{sec:two_sample_testing}

In this section, we focus on the non-parametric two-sample problem.
In \Cref{subsec:twosample_main}, we formally define the two-sample testing framework, explain how permutations or a wild bootstrap can be used to simulate the null, and present non-asymptotic level guarantees.
In \Cref{subsec:twosample_mmd_main} and \Cref{subsec:twosample_l2_main}, we present power guarantees in terms of MMD and $L^2$ Sobolev uniform separation rates, respectively, covering standard, efficent, differentially private and robust testing frameworks.
We refer the reader to \citet[Section 2.1]{schrab2025practical} for a detailed introduction to the Maximum Mean Discrepancy (MMD).

\mysubsection{Framework, bootstrap and level}
\label{subsec:twosample_main}

We first define the two-sample testing setting, we then present the permutation and wild bootstrap methods, which can be used to construct a test controlling the type I error non-asymptotically.

\paragraph{Two-sample testing.}\!\!\!\!\!\!\footnote{A more general framework is the one of credal two-sample testing, see \citet{chau2024credal} for details.}
Given independent i.i.d. samples $X_1,\dots,X_m$ from a distribution $P$, and i.i.d. samples $Y_1,\dots,Y_n$ from a distribution $Q$, the aim is to test whether the two distributions are equal, that is, $\Hcal_0\colon P = Q$, or not, \ie, $\Hcal_1\colon P \neq Q$.
Following the general hypothesis testing notation of \Cref{sec:def}, this corresponds to having $\Pcal$ as the space of all pairs of distributions, $\Pcal_0$ as $\{(P,Q)\in\Pcal : P=Q\}$, and $\Pcal_1$ as $\{(P,Q)\in\Pcal : P\neq Q\}$.
We use the notation $\Xbb_m \coloneqq (X_1,\dots,X_m)$, $\Ybb_n \coloneqq (Y_1,\dots,Y_n)$ and let $N=\min(m,n)$.

\paragraph{Exchangeability.}
Two-sample testing can be framed as testing for exchangeability (\Cref{sec:def}).
Permuted two samples are constructed by, first, combining the two original samples, permuting all the elements, and then splitting them again into two separate samples (of the original sample sizes).
Samples which come from the same distribution (\ie, null hypothesis) are exactly the ones which are exchangeable.
Indeed, it is clear that null samples, which are i.i.d., are exchangeable.
Under the alternative, both permuted samples can be seen as drawn i.i.d. from a mixture of the two distinct distributions, this shows that the original and permuted samples do not have the same distribution (\ie, samples are not exchangeable).
Hence, two-sample testing corresponds to testing for exchangeability, and the two-sample null hypothesis can be simulated using permutations.

\paragraph{Permutations.}
For notation purposes, let $Z_i = X_i$ for $i=1,\dots,n$ and $Z_i = Y_{i-m}$ for $i=n+1,\dots,m+n$.
As aforementioned, given a permutation $\pi$ of $\{1,\dots,m+n\}$, permuting the original samples $\Xm$ and $\Yn$ with respect to $\pi$ leads to the permuted samples $\Xm^\pi\coloneqq (Z_{\pi(1)},\dots,Z_{\pi(m)})$ and $\Yn^\pi\coloneqq (Z_{\pi(m+1)},\dots,Z_{\pi(m+n)})$.
Given any statistic function $T$, the test statistic is simply $T(\Xm,\Yn)$, and permuted statistics can be computed as $T(\Xm^\pi,\Yn^\pi)$ for various permutations $\pi$ randomly sampled.
A test can then be constructed using these permutations and can be performed efficiently, as explained in \Cref{subsec:level_hp} and \citet[Appendix C]{schrab2021mmd}.
This resulting test is well-calibrated with non-asymptotic level $\alpha$ as desired (\citealp[Lemma~1]{romano2005exact}, see also \citealp[Lemma 15]{kim2023differentially}).

\paragraph{Wild bootstrap.}
While the permutation approach presented above is applicable when using any type of MMD estimators (and more generally when using any other statistic). 
We now consider the wild bootstrap method presented in details in \Cref{subsec:level_hp}, which is specifically designed for (MMD) estimators expressed as one-sample second-order statistics (see \citealp[Section 3]{schrab2025practical}).
The wild bootstrapped statistics are computed as
\(
{|\Dcal|}^{-1}\sum_{(i,j) \in \Dcal} \varepsilon_i \varepsilon_j h(X_i, X_j)
\)
where $\varepsilon_1,\dots,\varepsilon_n$ are realisations of i.i.d.~Rademacher variables.
Using these allows to construct a test following the procedure of \Cref{subsec:level_hp}, the resulting test is guaranteed to control the type I error asymptotically \citep[Theorem 1]{chwialkowski2014wild}.
In the two-sample setting with $m=n$, the MMD estimator \citep[\emph{e.g.},][Equation 11]{schrab2025practical} admits as core function $\coremmd(x, x'; y, y') \coloneqq k(x,x') - k(x',y) - k(x,y') + k(y,y')$, which satisfies $\coremmd(x, y'; x', y) = -\coremmd(x, x'; y, y')$.
Leveraging this identity allows to prove that, when $m=n$, using a wild bootstrap is equivalent to using permutations which are only able to swap $X_i$ and $Y_i$ (or not) for $i=1,\dots,n$ \citep[Appendix B]{schrab2021mmd}.
This guarantees non-asymptotic type I error control of the wild bootstrap MMD test, which can be implemented efficiently \citep[Appendix C]{schrab2021mmd}.
See \Cref{subsec:level_hp} for computational details, as well as for a discussion on when to use each of the two bootstrapping methods.

\paragraph{Level.}
The permutation-based MMD, dpMMD and dcMMD tests (\Cref{sec:dp_dc}), as well as the wild bootstrap MMD test, all tightly control the probability of type I error by $\alpha$ at every sample size as desired
(\citealp[Proposition 1]{schrab2021mmd};
\citealp[Theorem 5]{kim2023differentially}; 
\citealp[Lemmas 1 and 4]{schrab2024robust}).
This non-asympotic level is preserved when using efficient estimators \citep[Proposition 1]{schrab2022efficient}, as well as when using adaptivity over kernels, either via pooling (properties of the permutation method, \citealp[Lemma~1]{romano2005exact}, combined with \citealp[Proposition 1]{schrab2021mmd}; see also discussion around \citealp[Theorem 1]{biggs2023mmd}) or via aggregation \citep[Proposition 8]{schrab2021mmd}.

\paragraph{Consistency (pointwise power).}
The MMD, dpMMD and dcMMD tests are all consistent in power: for large enough sample size, these two-sample tests can accurately detect any fixed alternative.
Consistency of these tests is guaranteed by \citet[Theorem 5]{kim2023differentially} and \citet[Lemmas 2 and 5]{schrab2024robust}.
Next, we derive stronger non-asymptotic power guarantees for which high power holds uniformly across alternatives shrinking with the sample sizes.

\paragraph{Kernel adaptivity.}
The MMD-based tests depend on the choice of kernel, which in practice greatly impacts the test power.
To solve this problem, kernel adaptivity can be performed either via aggregation (\Cref{sec:multiple_testing}) or via kernel pooling \citep[Section 4]{schrab2025practical}.
In practice, we recommend using the MMDAgg test \citep{schrab2021mmd} and the normalised MMDFuse test \citep{biggs2023mmd}.

\mysubsection{Uniform power against alternatives separated in MMD metric}
\label{subsec:twosample_mmd_main}

We present uniform separation rates in terms of the MMD metric, under which high power is guaranteed for the MMD-based two-sample test.
We consider the standard, efficient, private and robust testing frameworks.

\paragraph{Standard testing.}
For a fixed kernel $k$, the MMD test is powerful provided that (\Cref{subsec:kernel_proof} and \citealp[Theorem 7]{kim2023differentially})
\begin{equation}
\label{eq:mmd_kernel}
\mathrm{MMD}_k ~\gtrsim~ \sqrt{\frac{\max\!\big\{\!\log(1/\alpha), \, \log(1/\beta)\big\}}{N}}
\end{equation}
which is minimax optimal \citep[Theorem 8]{kim2023differentially}.
When using unnormalised kernel pooling \citep[Section 4]{schrab2025practical} with the mean pooling function, by leveraging the linearity of the discrepancy in the kernel \citep[Equation 75]{schrab2025practical}, we obtain the uniform separation rate 
(\Cref{subsec:pooled_proof})
\begin{equation}
\label{eq:mmd_kernel_mean}
\underset{k\in \Kcal}{\mathrm{mean}} ~ \mathrm{MMD}_k ~\gtrsim~ \sqrt{\frac{\max\!\big\{\!\log(1/\alpha), \, \log(1/\beta)\big\}}{N}}.
\end{equation}
The MMD test with unnormalised kernel pooling \citep[Section 4]{schrab2025practical} with kernel pooling function either max or fuse (with fusing parameter $\nu\geq \max(N,\log(|\Kcal|))$), achieves the uniform separation rate
(\Cref{subsec:pooled_proof} and \citealp[Theorems 2 and 3]{biggs2023mmd})
\begin{equation}
\label{eq:mmd_kernel_pooled}
\max_{k\in \Kcal}\,\mathrm{MMD}_k ~\gtrsim~ \sqrt{\frac{\max\!\big\{\!\log(1/\alpha), \, \log(1/\beta), \, \log(|\Kcal|)\big\}}{N}}
\end{equation}
where a typical choice is $|\Kcal|=\log(N)$ \citep[\emph{e.g.},][Corollary 10]{schrab2021mmd} leading to the rate $(N/\log\log N)^{-1/2}$ with an iterated logarithmic cost for adaptivity.
We note that for the fuse case, it is possible to get the fuse pooling function on the left hand side of \Cref{eq:mmd_kernel_pooled} instead of the maximum, but using the relation of \citet[Equation 78]{schrab2025practical} between both quantities, we can replace this fuse by the maximum provided that the fusing parameter $\nu$ is greater than both $N$ and $\log(|\Kcal|)$, which is almost always the case in practice.
In the remaining of this subsection, we present results only for fuse and max kernel pooling (variants of \Cref{eq:mmd_kernel_pooled}) without always mentioning the weak assumption $\nu\geq \max(N,\log(|\Kcal|))$, while results for kernel pooling with the mean function still hold with the exact same rate similarly to \Cref{eq:mmd_kernel_mean} but are not explicitly presented.

\paragraph{Efficient testing.}
The test using a block MMD B-statistic \citep[Equation 70]{schrab2025practical}, consisting of $B$ blocks, controls the type II error by $\beta$ if (\Cref{subsec:efficient_kernel_proof})
\begin{equation}
\label{eq:mmd_kernel_efficient}
\mathrm{MMD}_k ~\gtrsim~ \sqrt{\frac{B\max\!\big\{\!\log(1/\alpha), \, \log(1/\beta)\big\}}{N}}
\end{equation}
when using a fixed kernel $k$, and if 
(\Cref{subsec:pooled_proof})
\begin{equation}
\label{eq:mmd_kernel_efficient_pooled}
\max_{k\in \Kcal}\,\mathrm{MMD}_k ~\gtrsim~ \sqrt{\frac{B\max\!\big\{\!\log(1/\alpha), \, \log(1/\beta), \, \log(|\Kcal|)\big\}}{N}}
\end{equation}
when using a pooled unnormalised kernel collection $\Kcal$ \citep[Section 4]{schrab2025practical} with max or fuse pooling functions. 
It is possible to choose $\Kcal$ such that $|\Kcal|=\log(|\Dcal|/N)\approx\log(N/B)$ \citep[Theorem 2.ii]{schrab2022efficient} since the associated design $\Dcal$ is of size $B\lfloor N/B\rfloor^2\asymp N^2/B$.
When using a complete statistic (\ie, with $B=1$ block), the uniform separation rate achieved is the minimax one as in the standard testing setting above. 
As $B$ increases from $1$ to $N$, \Cref{eq:mmd_kernel_efficient} quantifies how the uniform separation rate deteriorates from being minimax to finally not even converging to zero. 
This highlights the trade-off between computational efficiency and power (speed of uniform separation rate).

\paragraph{Differentially private testing.}
The $(\varepsilon,\delta)$-differentially private dpMMD test
\citep[Algorithm 1 and Section 4]{kim2023differentially} 
achieves the uniform separation rate
(\Cref{subsec:dp_kernel_proof} and \citealp[Theorem 7]{kim2023differentially})
\begin{equation}
\label{eq:mmd_dp_ker}
\mathrm{MMD}_k ~\gtrsim~  \max \Biggl\{ \sqrt{\frac{\max\!\big\{\!\log(1/\alpha), \, 	
\log(1/\beta)\big\}}{N}}, \, \frac{\max\!\big\{\!\log(1/\alpha), \, \log(1/\beta)\big\}}{N\xi}  \Biggr\}
\end{equation}
which is minimax optimal \citep[Theorem 8]{kim2023differentially}, where 
$\xi=\varepsilon + \log\!\big({1}/({1-\delta})\big)$.
This holds for any fixed kernel $k$.
In the low privacy regime with $\xi \gtrsim \sqrt{{\max\!\big\{\!\log(1/\alpha),\log(1/\beta)\big\}}/{N}}$, differentially privacy comes for free as dpMMD achieves the non-DP minimax rate (\ie, first term).
In the high privacy regime with $\xi \lesssim \sqrt{{\max\!\big\{\!\log(1/\alpha),\log(1/\beta)\big\}}/{N}}$, the rate (\ie, second term) deteriorates gradually away from the non-DP minimax rate.

Using a kernel pooling method is possible but not straightforward in this differential privacy setting.
Recall that the dpMMD test injects privatisation noise in $B$ permuted statistics (and in the original statistic), to ensure differentially privacy a naive approach based on the composition theorem \citep[Lemma 2]{kim2023differentially} would scale the noise by $B$ and result in a powerless test, \citet[Lemma 4]{kim2023differentially} proves that scaling by a factor 2 (independent of $B$) is enough to guarantee differential privacy.
When using a pooling method with a collection of $|\Kcal|$ kernels, the naive approach would require the privatisation noise to scale with $|\Kcal|B$, a more refined approach would need to be derived.

\paragraph{Robust to data corruption testing.}
The dcMMD \citep[Algorithm 1 and Section 3]{schrab2024robust} and dpMMD \citep[Algorithm 2, Section 5, Appendix E]{schrab2024robust} tests with fixed kernel, designed to be robust against corruption of up to $r$ samples, are guaranteed to be powerful as soon as (\Cref{subsec:robust_kernel_proof} and \citealp[Theorems 1.i and 3]{schrab2024robust})
\begin{align}
\label{eq:mmd_kernel_robust}
\mathrm{MMD}_k
~\gtrsim~
\max \biggl\{ \sqrt{\frac{\max\{\log(1/\alpha),\log(1/\beta)\}}{N}}, \ \frac{r}{N} \biggr\} 
\end{align}
which is minimax optimal \citep[Theorem 1.ii]{schrab2024robust}.
Recall that the number $r$ of samples to be robust against is necessarily smaller or equal to $N$.
If $r \lesssim \sqrt{N \max\{\log(1/\alpha),\log(1/\beta)\}}$, then there is no price to pay for robustness as the robust tests achieve the minimax optimal rate of the standard non-robust testing framework.
As $r$ increases above $\sqrt{N \max\{\log(1/\alpha),\log(1/\beta)\}}$, the uniform separation rate becomes $r/N$ which is minimax optimal.
If $r=N$, then the rate no longer converges to zero, this means that the type II error cannot be controlled against any alternative, which indeed makes sense as in this setting all the samples can be corrupted and, hence, all information is lost.
Unnormalised kernel pooling, with either max or fuse pooling functions, leads to the uniform separation rate
\begin{align}
\label{eq:mmd_kernel_robust_pooled}
\max_{k\in \Kcal}\,\mathrm{MMD}_k
~\gtrsim~
\max \biggl\{ \sqrt{\frac{\max\{\log(1/\alpha),\log(1/\beta),\log(|\Kcal|)\}}{N}}, \ \frac{r}{N} \biggr\}.
\end{align}

\mysubsection{Uniform power against alternatives separated in L2 metric}
\label{subsec:twosample_l2_main}

We now report power guarantees in terms of uniform separation rates with respect to the $L^2$-norm of the difference in densities for the two-sample problem under the standard, efficient and private testing frameworks.
We consider translation-invariant kernels and impose a Sobolev smoothness requirement on the difference in densities $p-q$.

\paragraph{Standard testing.}
The uniform separation rate of the MMD test with an optimal bandwidth depending on the unknown Sobolev smoothness $s$ is (\Cref{subsec:l2_proof} and \citealp[Corollary 7]{schrab2021mmd})
\begin{align}
\label{eq:mmd_l2}
\|p-q\|_{L^2}
~\gtrsim~
\p{\frac{\log(1/\alpha)\log(1/\beta)}{N}}^{2s/(4s+d)}
\end{align}
which is minimax optimal \citep[Appendix D]{schrab2021mmd}.
If the difference in densities is not smooth (\ie, $s\to0$), the rate becomes constant and power cannot be guaranteed against any alternative.
If $p-q$ is very smooth (\ie, $s\to\infty$), then the uniform separation rates simply becomes of order $N^{-1/2}$.
We also remark that the rate deteriorates as the dimension $d$ increases.

Since the kernel bandwidth for the optimal test above depends on the unknown Sobolev smoothness, it cannot be implemented in practice.
By aggregating over various kernel bandwidths (all independent of the unknown Sobolev smoothness $s$) with multiple testing, we can construct an implementable test which achieves the uniform separation rate (\Cref{subsec:agg_l2_proof} and \citealp[Corollary 10]{schrab2021mmd})
\begin{align}
\label{eq:mmd_l2_agg}
\|p-q\|_{L^2}
~\gtrsim~
\p{\frac{\log(1/\alpha)\log(1/\beta)}{N/\log(\log(N))}}^{2s/(4s+d)}.
\end{align}
This aggregated multiple test is adaptive to the unknown Sobolev smoothness.
This comes only at the price of an iterated logarithmic term in the minimax rate, which is $\log|\Kcal|$ where $|\Kcal|\asymp \log N$ \citep[Corollary 10]{schrab2021mmd}.

\paragraph{Efficient testing.}
The efficient test, with an MMD estimator computable in time $\bigO{|\Dcal|}$ and with optimal kernel bandwidth (depending on the unknown Soboleve smoothness $s$), controls the type II error by $\beta$ when
(\Cref{subsec:eff_l2_proof} and \citealp[Theorem 2]{schrab2022efficient})
\begin{align}
\label{eq:efficient_l2}
\|p-q\|_{L^2}
~\gtrsim~
\p{\frac{\log(1/\alpha)\log(1/\beta)}{|\Dcal|/N}}^{2s/(4s+d)}.
\end{align}
When the complete statistics are used (\ie, $|\Dcal|\asymp N^2$) the rate is minimax optimal as in the standard testing framework above.
As the complexity $|\Dcal|$ decreases from $N^2$ to $N$, the uniform separation rate gradually deteriorates, until it no longer converges to zero.
This means that the result of \Cref{eq:efficient_l2} does not guarantee power for the linear-time MMD tests against any alternative.
However, due to the dependence of the bandwidth on the unknown Sobolev smoothness, this test cannot be implemented in practice.

Multiple testing via aggregation over a well-chosen collection of bandwidths \citep[Theorem 2.ii]{schrab2022efficient}, which is independent of the unknown Sobolev smoothness $s$, results in the uniform sepration rate
(\Cref{subsec:agg_l2_proof} and \citealp[Theorem 3]{schrab2022efficient})
\begin{align}
\label{eq:mmd_l2_eff_agg}
\|p-q\|_{L^2}
~\gtrsim~
\p{\frac{\log(1/\alpha)\log(1/\beta)}{\big(|\Dcal|/N\big)\big/\log\!\big(\!\log(|\Dcal|/N)\big)}}^{2s/(4s+d)}
\end{align}
which is the same as the rate of \Cref{eq:efficient_l2} up to an iterated logarithmic term $\log |\Kcal|$ where $|\Kcal|\asymp \log(|\Dcal|/N)$ \citep[Theorem 2.ii]{schrab2022efficient}.

\paragraph{Differentially private testing.}
The dpMMD test \citep[Algorithm 1 and Section 4]{kim2023differentially} which is $(\varepsilon,\delta)$-differentially private achieves different uniform separation rates depending on the privacy regime (\Cref{subsec:dp_l2_proof} and \citealp[Theorem 9]{kim2023differentially}).
Let $\xi=\varepsilon + \log\!\big({1}/({1-\delta})\big)$.
In the low privacy regime with $\xi \gtrsim N^{-(2s-d/2)/(4s+d)}$, power is guaranteed when
\begin{equation}
\label{eq:mmd_l2_dp_low}
\|p-q\|_{L^2}
~\gtrsim~
N^{-2s/(4s+d)}
\end{equation}
which is the non-DP minimax optimal rate \citep[Appendix D]{schrab2021mmd}. 
This means that, in this low privacy regime, differential privacy comes for free in the sense that there is no price to pay in the uniform separation rate for being differentially private.
In the mid privacy regime with $N^{-1/2} \lesssim\xi \lesssim N^{-(2s-d/2)/(4s+d)}$, the uniform separation rate is 
\begin{equation}
\label{eq:mmd_l2_dp_mid}
\|p-q\|_{L^2}
~\gtrsim~
( N^{{3}/{2}} \xi)^{-s/(2s+d)},
\end{equation}
and in the high privacy regime with $\xi\lesssim N^{-1/2}$ it is
\begin{equation}
\label{eq:mmd_l2_dp_high}
\|p-q\|_{L^2}
~\gtrsim~
\left(N\xi\right)^{-2s/(2s+d)}.
\end{equation}
These rates and privacy regimes are (not yet) guaranteed to be minimax optimal, deriving matching $L^2$ lower bounds for differentially private testing remains an open problem, which is left for future work.
These results are derived with a logarithmic dependence in $\alpha$, as mentioned in \Cref{subsec:dp_l2_proof}, we believe that obtaining a logarithmic dependence in $\beta$ is also possible as in the standard and efficient testing setting.

\mysection{Independence testing}
\label{sec:independence_testing}

We now consider the non-parametric independence testing problem.
In \Cref{subsec:ind_main}, we formally introduce the framework, its two null simulation methods (permutations and wild bootstrap) leading to a well-calibrated non-asymptotic test.
In \Cref{subsec:ind_hsic_main} and \Cref{subsec:ind_l2_main}, we provide power guarantees in terms of HSIC and $L^2$ Sobolev uniform separation rates, respectively, for independence testing under four different settings.
We refer the reader to \citet[Section 2.2]{schrab2025practical} for a detailed introduction to the Hilbert--Schmidt Independence Criterion (HSIC).

\mysubsection{Framework, bootstrap and level}
\label{subsec:ind_main}

First, we formalise the independence testing framework, we then explain how using either permutations or a wild bootstrap simulates the null and can be used to construct an independence test with the desired non-asymptotic level.

\paragraph{Independence testing framework.}
Given paired samples $(X_1,Y_1),\dots, (X_N,Y_N)$ drawn i.i.d. from a joint distribution $P_{XY}$, the aim is to test whether the first and second components of the pairs are independent, that is, $\Hcal_0\colon P_{XY} = P_X \otimes P_Y$, or dependent, \ie, $\Hcal_1\colon P_{XY} \neq P_X \otimes P_Y$.
Mirroring the notation of \Cref{sec:def} leads to defining $\Pcal$ as the space of all joint distributions, $\Pcal_0$ as the subspace of all products of marginals $\{P_{XY}\in\Pcal : P_{XY} = P_X \otimes P_Y\}$, and $\Pcal_1$ as $\{P_{XY}\in\Pcal : P_{XY} = P_X \otimes P_Y\}$.

\paragraph{Solving the independence problem with a two-sample test.}
Consider the independence testing problem where we are given paired samples $(X_1,Y_1),\dots, (X_N,Y_N)$ drawn i.i.d. from a joint distribution $P_{XY}$.
Note that the samples $X_1,\dots,X_N$ and $Y_1,\dots,Y_N$ naturally come from the marginals $P_X$ and $P_Y$, by pairing them randomly we can obtain samples from the product of the marginals $P_X \otimes P_Y$.
Since we are interested in testing for independence, \ie, whether the joint is equal to the product of marginals, it is possible to tackle this with a two-sample test with the first sample being sampled from the joint, and the second being sampled from the product of marginals.
However, to run a two-sample test, it is crucial that the two samples are independent from each other.
For this reason, we must split the paired samples into two separate paired samples (most likely of the same size), one which is left as is (\ie, sampled from the joint), and the other which is randomly shuffled (\ie, sampled from the product of marginals).
Running a two-sample test on these two samples then solves the independence problem.
Solving the independence problem in such a way using a two-sample test is, however, far from optimal due to the signal loss incurred as the potential dependence in the data is being ignored for half of it.
This justifies the need for metrics and tests specifically designed for the independence problem, such as the HSIC.
While the HSIC metric is equal to the MMD metric between the joint and the product of marginals (see paragraph `HSIC as an MMD' in \citealp[Section 2.2]{schrab2025practical}), its estimator as a fourth-order statistic is more complex than using an MMD estimator on paired data as described above, and exploits all the data and signal available \citep[Equation 21]{schrab2025practical}.

\paragraph{Solving the two-sample problem with an independence test.}
Consider the two-sample testing problem where we are given i.i.d.~samples $X_1,\dots,X_m$ from a distribution $P$, and i.i.d. samples $Y_1,\dots,Y_n$ from a distribution $Q$, all independent from each other, and we are interested in testing whether $P=Q$.
To frame this as an independence problem, we need to construct paired samples with dependence when $P\neq Q$, and no dependence when $P=Q$.
Consider the paired samples $(X_1,1),\dots, (X_m,1), (Y_1,-1),\dots, (Y_n,-1)$ where the second component of the pairs is an indicator of which sample the data point belongs to.
When $P=Q$, the two components of the pairs are independent, and when $P\neq Q$, the two components are dependent. 
Hence, performing an independence test on these paired samples solves the two-sample problem.
Computing the HSIC $V$-statistic on such paired samples with an indicator kernel for the labels is equivalent to computing a scaled MMD $V$-statistic on the original two-sample data (see \citealp[Equation 31]{schrab2025practical}), and the same holds for the HSIC and MMD metrics (see paragraph `MMD as an HSIC' in \citealp[Section 2.2]{schrab2025practical}).
If noise is required in the second component of the pairs, it can simply be added, for example, by adding i.i.d.~uniform noise on $[-\nicefrac{1}{2},\nicefrac{1}{2}]$, in which case the same equivalence results between MMD and HSIC hold provided a kernel $k^\Ycal(y,y') = \one(|y-y'| \leq 1)$ is used for the labels.

\paragraph{Exchangeability.}
Independence testing can also be framed as testing for exchangeability (\Cref{sec:def}).
To permute paired samples, only the elements in the second component of the pairs are permuted.
Independent paired samples (\ie, null hypothesis) are exactly the ones which are exchangeable.
Indeed, clearly, the independent paired samples are exchangeable since there is no dependence between the two components of the pairs.
Under the alternative, permuting the elements of the second component breaks the dependence, so the paired samples are not exchangeable as the original and permuted paired samples are not identically distributed. 
Therefore, independence testing corresponds to testing for exchangeability, and the independence null hypothesis can be simulated using permutations.

\paragraph{Permutations.}
As presented above, given a permutation $\pi$ of $\{1,\dots,N\}$, permuting the original paired samples $\Zn = ((X_i,Y_i))_{i=1}^N$ with respect to $\pi$ results in $((X_i,Y_{\pi(i)}))_{i=1}^N$.
For any statistic function $T$, the test statistic is $T(\Zn)$, and a permuted statistic can be computed as $T(\Zn^\pi)$ for any permutation $\pi$.
Following \Cref{subsec:level_hp}, a test can be constructed using randomly sampled permutations, it achieves type I error control at the prescribed level $\alpha$ non-asymptotically (\citealp[Lemma~1]{romano2005exact}, see also \citealp[Lemma 15]{kim2023differentially}).

\paragraph{Wild bootstrap.}
When using a one-sample second-order statistic (see \citealp[Section 3]{schrab2025practical}), the null can be simulated asymptotically with wild bootstrapped statistics 
\(
{|\Dcal|}^{-1}\sum_{(i,j) \in \Dcal} \varepsilon_i \varepsilon_j h(X_i, X_j)
\)
using i.i.d.~Rademacher variables $\varepsilon_1,\dots,\varepsilon_n$.
Relying on these, a test controlling the level asymptotically \citep[Theorem 1]{chwialkowski2014wild} can be constructed following the procedure described in \Cref{subsec:level_hp}.
When using the HSIC estimator of \citet[Equation 25]{schrab2025practical} with $\corehsick$ as in \citet[Equation 23]{schrab2025practical}, computing a wild bootstrapped statistic corresponds to computing a permuted statistic for some specific permutation allowed to swap $i$ with $i+N/2$ for $i=1,\dots,N/2$, for $N$ even (see \citealp[Appendix F.1]{schrab2022efficient} for details).
Leveraging this fact, non-asymptotic level control can be guaranteed for this HSIC wild bootstrap test.
Discussions regarding efficient implementations of such tests, as well as guidance on when to use each bootstrapping methods, are provided in \Cref{subsec:level_hp}.
See also \citet[Theorem 4]{pogodin2024practical} for wild bootstrap guarantees for quantities closely  related to HSIC. 

\paragraph{Level.}
The permutation-based HSIC, dpHSIC and dcHSIC tests (\Cref{sec:dp_dc}), as well as the wild bootstrap HSIC test, all control the probability of type I error at $\alpha$ at every sample size as desired
(\citealp[Proposition 1]{albert2019adaptive};
\citealp[Theorem 6]{kim2023differentially}; 
\citealp[Lemmas 1 and 4]{schrab2024robust}).
This non-asympotic level is preserved when using efficient estimators \citep[Proposition 1]{schrab2022efficient}, as well as when using adaptivity over kernels, either via pooling (properties of the permutation method, \citealp[Lemma~1]{romano2005exact}, combined with \citealt[Proposition 1]{albert2019adaptive}; see also discussion around \citealp[Theorem 1]{biggs2023mmd}) or via aggregation \citep[Section 3.1]{albert2019adaptive}.

\paragraph{Consistency (pointwise power).}
The HSIC, dpHSIC and dcHSIC tests all achieve pointwise power, that is, they are consistent in the sense that any fixed alternative can eventually be detected with power 1 for large enough sample size.
Consistency of these independence tests is guaranteed by \citet[Theorem 6]{kim2023differentially} and \citet[Lemmas 3 and 6]{schrab2024robust}.
Next, we derive non-asymptotic power guarantees which hold uniformly rather than pointwise, this enables to guarantee high power against alternatives which shrink with the sample size.

\paragraph{Kernel adaptivity.}
The power of the HSIC tests is greatly affected by the choice of the two kernels.
This issue of kernel selection can be addressed either via aggregation (\Cref{sec:multiple_testing}) or via kernel pooling \citep[Section 4]{schrab2025practical}. In practice, we recommend using the HSICAgg test \citep{albert2019adaptive} and the normalised HSICFuse test.

\mysubsection{Uniform power against alternatives separated in HSIC metric}
\label{subsec:ind_hsic_main}

We present power guarantees in terms of uniform separation rates in the HSIC metric for kernel independence tests in the standard, efficient, private and robust frameworks, with fixed and pooled kernels.

\paragraph{Standard testing.}
The HSIC-based independence test with fixed kernels $k$ and $\ell$ achieves high power if (\Cref{subsec:kernel_proof} and \citealp[Theorem 12]{kim2023differentially})
\begin{align}
\label{eq:std_ker_hsic}
\mathrm{HSIC}_{k,\ell}
~\gtrsim~
\sqrt{\frac{\max\!\big\{\!\log(1/\alpha), \, \log(1/\beta)\big\}}{N}}
\end{align}
which is minimax optimal \citep[Theorem 13]{kim2023differentially}.
Using mean kernel pooling \citep[Section 4]{schrab2025practical} results in the uniform separation rate
(\Cref{subsec:pooled_proof})
\begin{align}
\label{eq:std_ker_hsic_mean}
\underset{k\in\Kcal}{\mathrm{mean}} 
~ \underset{\ell\in\Lcal}{\mathrm{mean}} 
~ \mathrm{HSIC}_{k,\ell}
~\gtrsim~
\sqrt{\frac{\max\!\big\{\!\log(1/\alpha), \, \log(1/\beta)\big\}}{N}}.
\end{align}
Relying on unnormalised fuse/max kernel pooling for the HSIC test results in an additional logarithmic term in the size of the product kernel collection, that is (\Cref{subsec:pooled_proof})
\begin{align}
\label{eq:hsic_kernel_pooled}
\max_{k\in \Kcal}\, 
\max_{\ell\in \Lcal}\, 
\mathrm{HSIC}_{k,\ell}
~\gtrsim~
\sqrt{\frac{\max\!\big\{\!\log(1/\alpha), \, \log(1/\beta), \, \log(|\Kcal||\Lcal|)\big\}}{N}}
\end{align}
where the fusing parameter is assumed to satisfy $\nu\geq \max(N,\log(|\Kcal||\Lcal|))$, and where a typical choice is $|\Kcal|=|\Lcal|=\log(N)$ \citep[\emph{e.g.},][Corollary 10]{schrab2021mmd}.\footnote{The fuse extension from the MMD two-sample framework to the HSIC independence one has brilliantly been conducted by Ren (Michael) Guangyo as part of his UCL MSc Machine Learning Project supervised by Antonin Schrab and Arthur Gretton.}
As in the MMD case, for other testing constraints, we present only the results for max and fuse kernel pooling (not necessarily always mentioning the assumption on $\nu$), but the results for mean pooling hold with the same rate as for fixed kernel, similarly to \Cref{eq:std_ker_hsic_mean}.

\paragraph{Efficient testing.}
The efficient HSIC test, with block HSIC B-statistic \citep[Equation 70]{schrab2025practical} consisting of $B$ blocks, and fixed kernels $k$ and $\ell$, controls the type II error by $\beta$ when
(\Cref{subsec:efficient_kernel_proof})
\begin{align}
\label{eq:hsic_kernel_efficient}
\mathrm{HSIC}_{k,\ell}
~\gtrsim~
\sqrt{\frac{B\max\!\big\{\!\log(1/\alpha), \, \log(1/\beta)\big\}}{N}}
\end{align}
which achieves the standard minimax optimal rate of \Cref{eq:std_ker_hsic} when the complete U-statistic is used (\ie, $B=1$).
As the number of blocks $B$ is increased from $1$ to $N$, the uniform separation rate gradually slows down from $N^{-1/2}$ to $N^0$ (\ie, no longer converging to zero).
The unnormalised pooled fuse/max block HSIC test \citep[Section 4]{schrab2025practical} has uniform separation rate
(\Cref{subsec:pooled_proof})
\begin{align}
\label{eq:hsic_kernel_efficient_pooled}
\max_{k\in \Kcal}\, 
\max_{\ell\in \Lcal}\, 
\mathrm{HSIC}_{k,\ell}
~\gtrsim~
\sqrt{\frac{B\max\!\big\{\!\log(1/\alpha), \, \log(1/\beta), \, \log(|\Kcal||\Lcal|)\big\}}{N}}
\end{align}
where common collection choices lead to $|\Kcal|=|\Lcal|=\log(|\Dcal|/N)\approx\log(N/B)$ \citep[Theorem 2.ii]{schrab2022efficient} as the block design $\Dcal$ is of size $B\lfloor N/B\rfloor^2\asymp N^2/B$.

\paragraph{Differentially private testing.}
The $(\varepsilon,\delta)$-differentially private dpHSIC test \citep[Algorithm 1 and Appendix B.5]{kim2023differentially} with fixed kernels $k$ and $\ell$ is powerful when (\Cref{subsec:dp_kernel_proof} and \citealp[Theorem 12]{kim2023differentially})
\begin{align}
\label{eq:hsic_kernel_dp}
\mathrm{HSIC}_{k,\ell}
~\gtrsim~
\max \Biggl\{ \sqrt{\frac{\max\!\big\{\!\log(1/\alpha), \, \log(1/\beta)\big\}}{N}}, \, \frac{\max\!\big\{\!\log(1/\alpha), \, \log(1/\beta)\big\}}{N\xi}  \Biggr\}.
\end{align}
where $\xi=\varepsilon + \log\!\big({1}/({1-\delta})\big)$.
This uniform separate rate is minimax optimal in all privacy regimes \citep[Theorem 13]{kim2023differentially}.
The dpHSIC test achieves the same non-DP minimax optimal rate (independent of $\xi$) as the HSIC test in the low privacy regime with $\xi \gtrsim \sqrt{{\max\!\big\{\!\log(1/\alpha),\log(1/\beta)\big\}}/{N}}$.
In the high privacy regime with $\xi \lesssim \sqrt{{\max\!\big\{\!\log(1/\alpha),\log(1/\beta)\big\}}/{N}}$, the rate deteriorates and depends on $\xi$, but this is still the best attainable rate of any $(\varepsilon,\delta)$-differentially private test.

For differential privacy, kernel pooling is rendered difficult due to the fact that the privatisation noise would scale with the number of kernels, unless proved otherwise (see discussion below \Cref{eq:mmd_dp_ker} for details).

\paragraph{Robust to data corruption testing.}
Being robust against corruption of up to $r$ samples, the dcHSIC \citep[Algorithm 1 and Section 4]{schrab2024robust} and dpHSIC \citep[Algorithm 2, Section 5, Appendix E]{schrab2024robust} tests with fixed kernels $k$ and $\ell$ have uniform separation rate (\Cref{subsec:robust_kernel_proof} and \citealp[Theorem 2.i and 4]{schrab2024robust})
\begin{align}
\label{eq:hsic_kernel_robust}
\mathrm{HSIC}_{k,\ell}
~\gtrsim~
\max \biggl\{ \sqrt{\frac{\max\{\log(1/\alpha),\log(1/\beta)\}}{N}}, \ \frac{r}{N} \biggr\} 
\end{align}
which is minimax optimal \citep[Theorem 2.ii]{schrab2024robust}.
When the tests are required to be robust to only a few samples (\ie, $r\lesssim \sqrt{N\max\{\log(1/\alpha),\log(1/\beta)\}}$), they achieve the non-robust minimax optimal rate (\ie, first term).
When robustness is required against more samples (\ie, $r\gtrsim \sqrt{N\max\{\log(1/\alpha),\log(1/\beta)\}}$), the uniform separation rate is simply $r/N$ which is guaranteed to be the best rate achievable in this setting.
A test that is robust to the corruption of all of the data (\ie, $r=N$), is of course vacuous and does not achieve any power (\ie, rate does not converge to zero).
Relying on unnormalised fuse/max kernel pooling, power at least $1-\beta$ can be guaranteed when 
\begin{align}
\label{eq:hsic_kernel_robust_pooled}
\max_{k\in \Kcal}\, 
\max_{\ell\in \Lcal}\, 
\mathrm{HSIC}_{k,\ell}
~\gtrsim~
\max \biggl\{ \sqrt{\frac{\max\{\log(1/\alpha),\log(1/\beta),\log(|\Kcal||\Lcal|)\}}{N}}, \ \frac{r}{N} \biggr\}.
\end{align}

\mysubsection{Uniform power against alternatives separated in L2 metric}
\label{subsec:ind_l2_main}

For the independence kernel tests, we present uniform separation rates in terms of the $L^2$-norm of the difference between the joint $p_{xy}$ and the product of the marginals $p_x\otimes p_y$, with Sobolev regularity assumption on $p_{xy}-p_x\otimes p_y$. 
Translation-invariant kernels are used and their bandwiths varied.
Minimax optimal rates can be attained by using the optimal kernel bandwidth which depends on the unknown Sobolev smoothness (\ie, cannot be implementable).
Adaptivity over the unknown Sobolev smoothness $s$ can be achieved using aggregation with multiple testing (\Cref{sec:multiple_testing}) over the kernel bandwidths independently of $s$.

\paragraph{Standard testing.}
The HSIC test, with optimal kernel bandwidths depending on the unknown Sobolev smoothness $s$, controls the type II error by $\beta$ for alternatives satisfying (\citealp[Corollary 2]{albert2019adaptive} with theoretical quantiles, and \citealp[Proposition 8.7]{kim2020minimax} with permuted quantiles, see also \Cref{subsec:l2_proof})
\begin{align}
\label{eq:hsic_l2}
\|p_{xy}-p_x\otimes p_y\|_{L^2}
~\gtrsim~
\p{\frac{\log(1/\alpha)\log(1/\beta)}{N}}^{2s/(4s+d)}
\end{align}
which is minimax optimal \citep[Theorem 4]{albert2019adaptive}.
The rate deteriorates as the smoothness $s$ is reduced and as the dimension $d$ is increased.
The optimal rate for infinite smoothness $s\to\infty$ is of order $N^{-1/2}$.

To avoid the dependence on the unknown Sobolev smoothness $s$, we resort to aggregation (multiple testing) over a collection of pairs of bandwidths, of size $(\log N)^2$, independent of $s$. 
The resulting test achieves the minimax optimal rate up to an iterated logarithmic term
(\citealp[Corollary 3]{albert2019adaptive} with theoretical quantiles, and \citealp[Theorem 3]{schrab2022efficient} with estimated quantiles, see also \Cref{subsec:agg_l2_proof})
\begin{align}
\label{eq:hsic_l2_agg}
\|p_{xy}-p_x\otimes p_y\|_{L^2}
~\gtrsim~
\p{\frac{\log(1/\alpha)\log(1/\beta)}{N/\log(\log(N))}}^{2s/(4s+d)}.
\end{align}

\paragraph{Efficient testing.}
The test based on the efficient HSIC estimator with design $|\Dcal|$ and fixed kernels is powerful provided that (\Cref{subsec:eff_l2_proof} and \citealp[Theorem 1]{schrab2022efficient})
\begin{align}
\label{eq:hsic_l2_eff}
\|p_{xy}-p_x\otimes p_y\|_{L^2}
~\gtrsim~
\p{\frac{\log(1/\alpha)\log(1/\beta)}{|\Dcal|/N}}^{2s/(4s+d)}.
\end{align}
When $|\Dcal|\asymp N^2$, the rate matches the standard minimax optimal rate.
As $|\Dcal|$ decreases, it deteriorates until it no longer converges for linear tests with $|\Dcal|\asymp N$.
This uniform separation rate holds assuming that optimal kernel bandwidths are used, as these depend on the unknown Sobolev smoothness $s$, this test cannot be implemented in practice.
To overcome this issue, one can use aggregation over the bandwidths (multiple testing), losing the unwanted dependence on $s$.
This results in the same power guarantee up to an iterated logarithmic term
(\Cref{subsec:agg_l2_proof} and \citealp[Theorem 2]{schrab2022efficient})
\begin{align}
\label{eq:hsic_l2_eff_agg}
\|p_{xy}-p_x\otimes p_y\|_{L^2}
~\gtrsim~
\p{\frac{\log(1/\alpha)\log(1/\beta)}{\big(|\Dcal|/N\big)\big/\log\!\big(\!\log(|\Dcal|/N)\big)}}^{2s/(4s+d)}.
\end{align}

\paragraph{Differentially private testing.}
The differentially private dpHSIC test
\citep[Algorithm 1 and Appendix B.5]{kim2023differentially} achieves different uniform separation rates depending on the value of $\xi=\varepsilon + \log\!\big({1}/({1-\delta})\big)$ compared to rates in $N$ (\Cref{subsec:dp_l2_proof} and \citealp[Theorem 14]{kim2023differentially}).
This creates three different privacy regimes.
In the low privacy regime with $\xi\gtrsim N^{-(2s-d/2)/(4s+d)}$, dpHSIC achieves the non-DP minimax rate
\begin{equation}
\label{eq:hsic_l2_dp_low}
\|p_{xy}-p_x\otimes p_y\|_{L^2}
~\gtrsim~
N^{-2s/(4s+d)},
\end{equation}
so privacy comes for free in this regime.
In the mid privacy regime with $N^{-1/2} \lesssim\xi \lesssim N^{-(2s-d/2)/(4s+d)}$, dpHSIC is powerful provided that
\begin{equation}
\label{eq:hsic_l2_dp_mid}
\|p_{xy}-p_x\otimes p_y\|_{L^2}
~\gtrsim~
( N^{3/2} \xi)^{-s/(2s+d)}.
\end{equation}
In the high privacy regime with $\xi \lesssim N^{-1/2}$, the uniform separation rate is
\begin{equation}
\label{eq:hsic_l2_dp_high}
\|p_{xy}-p_x\otimes p_y\|_{L^2}
~\gtrsim~
\left(N\xi\right)^{-2s/(2s+d)}.
\end{equation}
As in the two-sample case, these rates logarithmically depend on $\alpha$ and we believe that a logarithmic dependence in $\beta$ can also be obtained (\Cref{subsec:dp_l2_proof}).
Deriving matching lower bounds for $L^2$ separation under differential privacy constraint is an open problem, left for future work.

\mysection{Goodness-of-fit testing}
\label{sec:gof_testing}

Finally, we consider a third framework: non-parametric goodness-of-fit testing.
In \Cref{subsec:gof_main}, we define the goodness-of-fit testing framework and its associated wild-bootstrap which allows to compute a quantile and, hence, also to construct a well-calibrated test.
In \Cref{subsec:gof_ksd_main} and \Cref{subsec:gof_l2_main}, we provide power guarantees in terms of KSD and $L^2$ Sobolev uniform separation rates, respectively, under standard, efficiency, privacy and robustness constraints.
We refer the reader to \citet[Section 2.3]{schrab2025practical} for a detailed introduction to the Kernel Stein Discrepancy (KSD).

\mysubsection{Framework, bootstrap and level}
\label{subsec:gof_main}

We define the goodness-of-fit testing problem and construct a KSD test using a wild bootstrap which allows control of the type I error.

\paragraph{Goodness-of-fit testing framework.}
Given access to a model distribution $P$ and to i.i.d.~samples $X_1,\dots,X_N$ from a distribution $Q$, the aim is to test whether the two distributions are equal, that is, $\Hcal_0\colon P = Q$, or not, \ie, $\Hcal_1\colon P \neq Q$.
This goodness-of-fit problem is sometimes referred to as one-sample testing.
As in the two-sample case, the general setting of \Cref{sec:def} is covered by having $\Pcal$ as the space of all pairs of distributions, $\Pcal_0$ as $\{(P,Q)\in\Pcal : P=Q\}$, and $\Pcal_1$ as $\{(P,Q)\in\Pcal : P\neq Q\}$.
We use the notation $\Xbb_N \coloneqq (X_1,\dots,X_N)$.

\paragraph{Access to the model distribution.}
The type of access to the model distribution $P$ can differ depending on the testing setting.
If a simulator allowing to sample from $P$ is available, then the goodness-of-fit problem essentially reduces to the two-sample problem where as many samples as desired can be requested from the model distribution.
Being able to sample from the model, essentially simulating the null hypothesis, is closely related to the notion of parametric bootstrap \citep{stute1993bootstrap}.
In other cases, it might not be possible to have access to a simulator, but we can have some knowledge about the model distribution itself.
For example, it can sometimes be possible to compute the kernel expectations under the model in closed form, in which case a test can be constructed using the one-sample MMD plug-in estimator $\mathrm{MMD}_k^2(P,\widehat Q)$ (with the MMD as expressed in \citealp[Equation 6]{schrab2025practical}) which is equal to
\begin{equation}
	\label{eq:one_sample_mmd}
	\frac{1}{N^2} \sum_{1\leq i, j \leq N} k(X_i,X_{j})
	- \frac{2}{N} \sum_{i=1}^N  \mathbb{E}_{P}\big[k(X_i,Y)\big]
	+ \mathbb{E}_{P,P}\big[k(Y,Y')\big].
\end{equation}
This can equivalently be viewed as a KSD estimator with the simple Stein operator of \citet[Equation 58]{schrab2025practical}.
Another setting is the one where the density $p$ (with respect to the Lebesgue measure) of the model distribution is known.
However, this requires the model normalisation to known.
In order to allow for unnormalised models (\eg, energy-based models, normalising flows), we can instead assume that only the score $\nabla \log p$ of the model is known and accessible.
This is the most general goodness-of-fit setting, and the one we focus on in this work.

\paragraph{Solving the goodness-of-fit problem with a two-sample test.}
Consider the goodness-of-fit problem where we are given i.i.d.~samples $X_1,\dots,X_N$ from $Q$ and a model distribution $P$ from which we can draw samples (\ie, via a simulator).
Then, a two-sample test can be performed to solve this goodness-of-fit problem.
If an MMD estimator is computed with a Stein kernel \citep[Equation 43]{schrab2025practical}, as the model sample size grows to infinity, the MMD estimator converges to the KSD estimator.
Indeed, the terms involving the model samples in the MMD estimator approximate kernel expectations under the model, which are zero by Stein's identity, the term involving only the samples from $Q$ is the KSD estimator itself.
However, this only works when one is able to sample from the model, which is not the setting we consider here (\ie, access only to the model score function).

\paragraph{Solving the two-sample problem with a goodness-of-fit test.}
Consider the two-sample testing problem where we are given i.i.d.~samples $X_1,\dots,X_m$ from a distribution $P$, and i.i.d.~samples $Y_1,\dots,Y_n$ from a distribution $Q$, all independent from each other, and we are interested in testing whether $P=Q$.
As the model distribution, we can consider the empirical distribution $\widehat P$ (uniform distribution on the samples).
In this setting, we are then able to compute expectations under the model (empirical) distribution.
Running a goodness-of-fit test specialised for this setting can then be used to solve the original two-sample problem.
For example, we can use the one-sample MMD estimator of \Cref{eq:one_sample_mmd} (corresponding to the KSD estimator with the simple Stein operator of \citealp[Equation 58]{schrab2025practical}) with the model empirical distribution $\widehat P$.
It clear that computing the expectations with respect to $\widehat P$ in this estimator then leads to the usual MMD estimator.

\paragraph{Wild bootstrap.}
To construct a KSD test following the procedure of \Cref{subsec:level_hp}, we need a method to simulate the null when using a KSD one-sample second-order statistic (see \citealp[Section 3]{schrab2025practical}).
For this, we rely on the wild bootstrapped statistics (see \Cref{subsec:level_hp} for details) which can be computed as
\(
{|\Dcal|}^{-1}\sum_{(i,j) \in \Dcal} \varepsilon_i \varepsilon_j h_P(X_i, X_j)
\)
where $\varepsilon_1,\dots,\varepsilon_n$ are realisations of i.i.d.~Rademacher variables, and $h_P$ is the Stein kernel as defined in \citet[Equation 43]{schrab2025practical}.
The KSD test resulting from the test construction of \Cref{subsec:level_hp} with the wild bootstrap is then guaranteed to control the type I error asymptotically (\citealp[Theorem 1]{chwialkowski2014wild}).
The wild bootstrapped statistics can all be computed efficiently as outlined in \Cref{subsec:level_hp}. 

\paragraph{Level.}
The KSD test relying on a quantile computed via wild bootstrap controls the type I error at the desired level $\alpha$ asymptotically 
(\citealp[Proposition 3.2]{chwialkowski2016kernel}, and \citealp[Theorem 4.3]{liu2016kernelized}).
The goodness-of-fit test based on an efficient KSD estimator also achieves this asymptotic level control \citep[Proposition 1]{schrab2022efficient}.
Relying on kernel adaptation via aggregation with multiple testing also preserves the asymptotic type I error control \citep[Proposition 3.2]{schrab2022ksd}.
In the next paragraph, we explain how the level control can also be guaranteed when using kernel pooling for adaptivity.

\paragraph{Level: kernel pooling.}
In the two-sample and independence cases, the validity of the tests using kernel pooling with a wild bootstrap is guaranteed due to the correspondence of the wild bootstrap to a subgroup of permutations (\citealp[Appendix B]{schrab2021mmd} and \citealp[Appendix F.1]{schrab2022efficient}). 
Hence, the permutation validity guarantees (holding for any statistic) can be leveraged using a pooled statistic, even for the wild bootstrap.
However, in the goodness-of-fit setting this correspondence is broken and permutations cannot be used.
In order to prove that the KSD test using kernel pooling controls the level as desired, we need to show that the asymptotic distributions of the pooled KSD estimator and pooled wild bootstrap estimator are matching.
By linearity of the wild bootstrap statistic with respect to the kernel, and the fact that the original and wild bootstrap KSD estimators with the mean kernel have the same asymptotic distributions \citep[Theorem 1]{chwialkowski2014wild}, Cramér–Wold Theorem guarantees that the joint distribution of the wild bootstrap KSD statistics for all kernels in the collection is asymptotically the same as the joint distribution of the KSD statistics for all the kernels.
Then, the continuous mapping theorem (with continuous mean/max/fuse functions) guarantees that the pooled KSD wild bootstrap and original estimators have the same asympotitic distribution, and, hence,
asymptotic type I error control is guaranteed for KSD tests using a kernel pooling method.

\paragraph{Consistency (pointwise power).}
The KSD test is known to be consistent achieving pointwise power (\citealp[Proposition 3.2,]{chwialkowski2016kernel} and \citealp[Proposition 4.2]{liu2016kernelized}), that is, for any fixed alternative, the power of the KSD test converges to 1 asymptotically.
In \Cref{subsec:gof_ksd_main,subsec:gof_l2_main}, we present asymptotic power guarantees in a setting in which alternatives depend on the sample size and shrink towards the null as it increases.

\paragraph{Differential privacy, robustness, and permutations.}
The permutation-based privatisation and robustisation procedures of \citet[Algorithm 1]{kim2023differentially} and \citet[Algorithm 1]{schrab2024robust} only work in frameworks where testing the null corresponds exactly to testing exchangeability (see \Cref{sec:def}).
As previously mentioned, the two-sample and independence testing frameworks satisfy this property.
However, this is not the case for goodness-of-fit testing.
Indeed, since we have access to only one sample (and to the model itself), we are not able to permute samples across distributions as in the two-sample case, and permuting the points within one sample has no effect.
So, the goodness-of-fit problem cannot be framed as testing for exchangeability. 
This is the reason why there are no permutation method in this setting, and why the type I error control only holds asymptotically.
This also explains why the procedures of \citet[Algorithm 1]{kim2023differentially} and \citet[Algorithm 1]{schrab2024robust} cannot be used to construct private and robust KSD tests: a task which remains an open problem.

\paragraph{Kernel adaptivity.}
The choice of base kernel and its bandwidth leads to drastically different Stein kernels, the role of the bandwidth is even more crucial in the goodness-of-fit setting as it affects the derivates of the base kernel which appear in the expression of the Stein kernel.
Hence, this choice greatly impacts the power of the KSD test.
This kernel selection problem can be solved with the aggregation method (\Cref{sec:multiple_testing}) or with kernel pooling \citep[Section 4]{schrab2025practical}.
We recommend using these two adaptive procedures, leading to the KSDAgg test \citep[Algorithm 1]{schrab2022ksd} and the normalised KSDFuse test.

\mysubsection{Uniform power against alternatives separated in KSD metric}
\label{subsec:gof_ksd_main}

For the standard and efficient testing frameworks, we provide uniform separation rates in terms of the KSD metric, with the assumption on the kernel $k$ that its associated Stein kernel is bounded \citep[\eg,][Theorem 4.8]{barp2022targeted}, which could potentially be relaxed to a sub-Gaussian assumption leveraging results from \citet{kalinke2024nystrom}.
The methods of \citet[Algorithm 1]{kim2023differentially} and \citet[Algorithm 1]{schrab2024robust} to construct private and robust tests are based on a non-asymptotic permutation approach and, hence, do not apply to the KSD goodness-of-fit setting.

\paragraph{Standard testing.}
The KSD test is guaranteed to have power at least $1-\beta$ against all alternatives separated as
(\Cref{subsec:kernel_proof})
\begin{align}
\label{eq:ksd_ker_std}
\mathrm{KSD}_{k}
~\gtrsim~
\sqrt{\frac{\max\!\big\{\!\log(1/\alpha), \, \log(1/\beta)\big\}}{N}}
\end{align}
which is of order $N^{-1/2}$.
Using any unnormalised mean kernel pooling \citep[Section 4]{schrab2025practical} leads to 
(\Cref{subsec:pooled_proof})
\begin{align}
\label{eq:ksd_kernel_pooled_mean}
\underset{k\in\Kcal}{\mathrm{mean}} ~ 
\mathrm{KSD}_{k}
~\gtrsim~
\sqrt{\frac{\max\!\big\{\!\log(1/\alpha), \, \log(1/\beta) \big\}}{N}},
\end{align}
while using unnormalised fuse/max kernel pooling leads to
(\Cref{subsec:pooled_proof})
\begin{align}
\label{eq:ksd_kernel_pooled}
\max_{k\in \Kcal}\,
\mathrm{KSD}_{k}
~\gtrsim~
\sqrt{\frac{\max\!\big\{\!\log(1/\alpha), \, \log(1/\beta), \, \log(|\Kcal|)\big\}}{N}},
\end{align}
which, for a collection of kernels of size $|\Kcal|=\log N$ (\eg, \citealp[Theorem 3.5.ii]{schrab2022ksd}), corresponds to a rate of order $(N/\log\log N)^{-1/2}$.
Here, and throughout this section, the fusing parameter $\nu$ is assumed to be greater than $N$ and $\log(|\Kcal|)$.
In the following, we present results for fuse/max kernel pooling, while mean pooling results can be derived in a similar way to \Cref{eq:ksd_kernel_pooled_mean} achieving the same rate as for fixed kernel.

\paragraph{Efficient testing.}
The efficient goodness-of-fit test, based on the block KSD B-statistic \citep[Equation 70]{schrab2025practical} consisting of $B$ blocks, with fixed kernel $k$, controls the type II error by $\beta$ provided that
(\Cref{subsec:efficient_kernel_proof})
\begin{align}
\label{eq:ksd_kernel_efficient}
\mathrm{KSD}_{k}
~\gtrsim~
\sqrt{\frac{B\max\!\big\{\!\log(1/\alpha), \, \log(1/\beta)\big\}}{N}}.
\end{align}
If only one block is used (\ie, $B=1$ corresponding to the complete U-statistic), then the same rate $N^{-1/2}$ as in \Cref{eq:ksd_ker_std} is achieved.
As the number of blocks $B$ increases, the rate $(B/N)^{1/2}$ gets slower until it no longer converges to zero for $B\asymp N$.
The unnormalised fuse/max pooled block KSD statistic over a collection $\Kcal$ of kernels 
(often of size $|\Kcal|=\log(|\Dcal|/N)\approx\log(N/B)$ as in \citet[Theorem 2.ii]{schrab2022efficient} with $|\Dcal|=B\lfloor N/B\rfloor^2\asymp N^2/B$)
achieves the same uniform separation rate with an additional $\sqrt{\log |\Kcal|}$ term, that is
(\Cref{subsec:pooled_proof})
\begin{align}
\label{eq:ksd_kernel_efficient_pooled}
\max_{k\in \Kcal}\,
\mathrm{KSD}_{k}
~\gtrsim~
\sqrt{\frac{B\max\!\big\{\!\log(1/\alpha), \, \log(1/\beta), \, \log(|\Kcal|)\big\}}{N}}.
\end{align}

\mysubsection{Uniform power against alternatives separated in L2 metric}
\label{subsec:gof_l2_main}

We present power guarantees for the KSD test against alternatives separated in terms of the $L^2$-norm of the difference in scores multiplied by the data density, that is $\big\|\big(\nabla \log p - \nabla \log q\big)\,q\,\big\|_{L^2}$.
This score-based metric is perfectly suited for the KSD framework as it corresponds exactly to the quantity considered by \citet[Proposition 3.3]{liu2016kernelized} who introduced the KSD with \citet{chwialkowski2016kernel}.
Moreover, a Sobolev regularity assumption on $\big(\nabla \log p - \nabla \log q\big)\,q$ is made.
Intuitively, this imposes some smoothness restrictions both on the difference in scores and on the data density itself.

\paragraph{Standard testing.}
The KSD test with some specific kernel bandwidth depending on the unknown Sobolev smoothness is guaranteed to be powerful when
(\Cref{subsec:l2_proof})
\begin{align}
\label{eq:ksd_pow_l2}
\big\|\big(\nabla \log p - \nabla \log q\big)\,q\,\big\|_{L^2}
~\gtrsim~
\p{\frac{\log(1/\alpha)\log(1/\beta)}{\sqrt N}}^{2s/(4s+5d)}
\end{align}
which is weaker than the two-sample and independence minimax optimal rates.
The rate is derived in Appendix \ref{subsec:ksd_l2_proof_} and we believe it can be improved.
With stronger smoothness requirements (\ie, $s\to\infty$), the rate becomes $N^{-1/4}$.
If the smoothness is very weak (\ie, $s\to0$) or the dimension is very large (\ie, $d\to\infty$), then the rate $N^0$ no longer converges to zero and power cannot be guaranteed.

To avoid the dependence on the unknown Sobolev smoothness $s$ for practical uses, one can rely on aggregating over a collection of kernel bandwidths independent of $s$ (multiple testing), to obtain the uniform separation rate
(\Cref{subsec:agg_l2_proof})
\begin{align}
\label{eq:ksd_l2_agg}
\big\|\big(\nabla \log p - \nabla \log q\big)\,q\,\big\|_{L^2}
~\gtrsim~
\p{\frac{\log(1/\alpha)\log(1/\beta)}{\sqrt{N/\log(\log(N))}}}^{2s/(4s+5d)}.
\end{align}

\paragraph{Efficient testing.}
Using a KSD estimator with design $\Dcal$, the goodness-of-fit test with kernel bandwidth depending on $s$, the unknown Sobolev smoothness, controls the type II error by $\beta$ for alternatives satisfying
(\Cref{subsec:eff_l2_proof})
\begin{align}
\label{eq:ksd_l2_eff}
\big\|\big(\nabla \log p - \nabla \log q\big)\,q\,\big\|_{L^2}
~\gtrsim~
\p{\frac{\log(1/\alpha)\log(1/\beta)}{\sqrt{|\Dcal|/N}}}^{2s/(4s+5d)}.
\end{align}
Again, if a complete statistic is used (\ie, $|\Dcal|\asymp N^2$), then this is the same rate as \Cref{eq:ksd_pow_l2} in the standard framework.
There is a trade-off between efficiency and power in terms of speed of the uniform separation rate: when the complexity $|\Dcal|$ decreases, the set of detectable alternatives shrinks, until it is finally empty for linear tests with $|\Dcal|\asymp N$.

When relying on aggregation  over various kernel bandwidths independently of the unknown Sobolev smoothness (multiple testing), the resulting test is powerful provided that
(\Cref{subsec:agg_l2_proof})
\begin{align}
\label{eq:ksd_l2_eff_agg}
\big\|\big(\nabla \log p - \nabla \log q\big)\,q\,\big\|_{L^2}
~\gtrsim~
\p{\frac{\log(1/\alpha)\log(1/\beta)}{\sqrt{\big(|\Dcal|/N\big)\big/\log\!\big(\!\log(|\Dcal|/N)\big)}}}^{2s/(4s+5d)}
\end{align}
which is the same rate with an additional iterated logarithmic term.

\mysection{Open problems for future work}
\label{sec:future_work}

The above results provide an almost complete overview of the power guarantees of kernel-based tests in the two-sample, independence, and goodness-of-fit settings.
However, there still remains some open questions and directions for future work that we outline below.

\begin{enumerate}
\item $L^2$ separation upper bound for the HSIC test which also holds for smoothness $s\in\big(0, (d_{\mathcal{X}}+d_{\mathcal{Y}})/4\big)$ (\eg, \Cref{eq:hsic_l2}).
\item $L^2$ separation lower bounds under differential privacy constraint (\eg, \Cref{eq:mmd_l2_dp_low,eq:mmd_l2_dp_mid,eq:mmd_l2_dp_high,eq:hsic_l2_dp_low,eq:hsic_l2_dp_mid,eq:hsic_l2_dp_high}).
\item $L^2$ and kernel separation lower bounds for efficient testing under computational complexity constraint (\eg, \Cref{eq:mmd_kernel_efficient,eq:hsic_kernel_efficient,eq:ksd_kernel_efficient,eq:efficient_l2,eq:hsic_l2_eff,eq:ksd_l2_eff}).
\item $L^2$ separation upper and lower bounds under the `robustness to data corruption' constraint (both with optimal unknown kernel and aggregation).
\item $L^2$ separation lower bounds, and improved upper bounds, for goodness-of-fit testing (\eg, \Cref{eq:ksd_pow_l2}).
\item Kernel separation rates for normalised pooling (\eg, \Cref{eq:mmd_kernel_pooled,eq:hsic_kernel_pooled,eq:ksd_kernel_pooled} for unnormalised, \citealp[Theorem 3]{biggs2023mmd} for normalised).
\item Kernel pooling and aggregation procedure for differentially private tests with noise scaled independently of the number of kernels (see discussion below \Cref{eq:mmd_dp_ker}).
\item KSD private and robust test constructions, and uniform separation power guarantees (see discussions in \Cref{subsec:gof_main}).
\end{enumerate}

Solving these problems would really provide a complete power analysis of MMD, HSIC and KSD kernel-based tests in the two-sample, independence, and goodness-of-fit settings, under various testing constraints.\footnote{If you solve any of these problems, please let me know!}

\appendix
\renewcommand{\thesection}{A}
\mysection{Proof sketches}
\label{sec:proof_sketches}

The aim of this section is to provide intuition behind the proofs of the results presented above. We highlight the main proof steps while simplifying the tedious computations (\eg, constant factors) for ease of presentation. 
The full proofs are provided in the following chapters and are referenced here.
We present the structure of this section here.
\begin{itemize}
\item 
\Cref{subsec:kernel_proof}:
{Proof sketch of kernel separation} (V/U-statistics).
\item 
\Cref{subsec:dp_kernel_proof}:
{Proof sketch of differentially private kernel separation} (V-statistics).
\item 
\Cref{subsec:robust_kernel_proof}:
{Proof sketch of robust kernel separation} (V-statistics).
\item 
\Cref{subsec:pooled_proof}:
{Proof sketch of pooled kernel separation} (V/U-statistics).
\item 
\Cref{subsec:efficient_kernel_proof}:
{Proof sketch of efficient kernel separation} (B-statistics).
\item 
\Cref{subsec:pooled_efficient_proof}:
{Proof sketch of pooled efficient kernel separation} (B-statistics).
\item 
\Cref{subsec:l2_proof}:
{Proof sketch of $L^2$ separation} (U-statistics).
\item 
\Cref{subsec:eff_l2_proof}:
{Proof sketch of efficient $L^2$ separation} (incomplete U-statistics).
\item 
\Cref{subsec:agg_l2_proof}:
{Proof sketch of aggregated $L^2$ separation} (U-statistics).
\item 
\Cref{subsec:agg_eff_l2_proof}:
{Proof sketch of aggregated efficient $L^2$ separation} (incomplete U-statistics).
\item 
\Cref{subsec:dp_l2_proof}:
{Proof sketch of differentially private $L^2$ separation} (U-statistics).
\end{itemize}

\noindent
In general, the kernel separation results
(\Cref{subsec:kernel_proof,subsec:dp_kernel_proof,subsec:efficient_kernel_proof,subsec:robust_kernel_proof,subsec:pooled_proof,subsec:pooled_efficient_proof}) hold naturally when using the square-rooted V-statistic $\sqrt{V_k}$ with kernel $k$,
and the $L^2$ separation results
(\Cref{subsec:l2_proof,subsec:eff_l2_proof,subsec:agg_l2_proof,subsec:agg_eff_l2_proof,subsec:dp_l2_proof}) hold naturally when using the unbiased U-statistic $U_k$ with kernel $k$.
The classical kernel separation result (\Cref{subsec:kernel_proof}) can also be proved for U-statistics by leveraging their relation to V-statistics, and the efficient kernel separation result (\Cref{subsec:efficient_kernel_proof}) holds only for block B-statistics (not other incomplete U-statistics) because these can be related to a non-negative block incomplete V-statistic version.
Meanwhile, the efficient (aggregated) $L^2$ separation results (\Cref{subsec:eff_l2_proof,subsec:agg_eff_l2_proof}) hold for any incomplete U-statistics.

\subsection{Proof sketch of kernel separation}
\label{subsec:kernel_proof}

We detail the proof structure of the kernel separation results:
\Cref{eq:mmd_kernel} proved in \citet[Theorem 7]{kim2023differentially}, 
\Cref{eq:std_ker_hsic} proved in \citet[Theorem 12]{kim2023differentially},
and \Cref{eq:ksd_ker_std} proved here. 

\paragraph{V-statistic.}
Consider the statistic $\sqrt{V_0}$, where $V_0\coloneqq V_k$ is a V-statistic (\eg, \citealp[Equations 7, 19 and 52]{kim2023differentially}) for some fixed kernel $k$, $\sqrt{V_0}$ is an estimate of some kernel discrepancy $\kdisc$ (\eg, MMD, HSIC, KSD) for which $\kdisc=0$ characterises the null.
We also consider bootstrapped statistics $\sqrt{V_1},\dots,\sqrt{V_B}$ (either via permutations or wild bootstrap, \Cref{subsec:level_hp}), and the $(1-\alpha)$-quantile $q_{1-\alpha}$ of $\sqrt{V_0},\dots,\sqrt{V_B}$.
The aim is to bound the probability of type II error $\PP(T_0 \leq q_{1-\alpha})$ by $\beta$ provided that the discrepancy $\kdisc$ is greater than some rate to be determined.
For this, we need two results: some exponential concentration bounds for the original statistic and for the bootstrapped statistic.

The first one is a concentration inequality for the quantity $\sqrt{V_0}$ estimating $\kdisc$ of the form
\begin{align}
\mP\Big( \big|\sqrt{V_0} - \kdisc \big| > \widetilde{C}\, N^{-1/2} + t \Big) ~\leq~ \exp(-C \,t^2 N )
\end{align}
for all $t>0$, which leads to 
\begin{align}
\label{eq:concentration_sqrt_V}
\big|\sqrt{V_0} - \kdisc \big| ~\leq~ C_1 \sqrt{\frac{1}{N} \log \biggl(\frac{1}{\beta}\biggr)}
\end{align}
with probability at least $1-\beta/2$.
Such statistic concentration results can be found in \citet[Lemma 13]{kim2023differentially} for MMD, and in \citet[Lemma 14]{kim2023differentially} for HSIC. The KSD V-statistic case can be derived from the KSD U-statistic case (holding by Hoeffding's inequality, \citealp[Equation 5.7]{hoeffding63}) using the two facts that
$U_{\mathrm{KSD}} (N-1)/N \leq V_{\mathrm{KSD}} \leq K/N + U_{\mathrm{KSD}} (N-1)/N$, where $K$ is a bound on the kernel, and that
$
\big|V_0-\kdisc^2\big|
= 
\big|\sqrt{V_0} - \kdisc\big| \big|\sqrt{V_0} + \kdisc\big| 
\leq
2K \big|\sqrt{V_0} - \kdisc\big|
$.
The U-statistics and V-statistics for MMD and HSIC can also be expressed in terms of each other \citep[Appendix E.13 and Lemma 22]{kim2023differentially}.

The second concentration inequality is on the bootstrapped statistic $\sqrt{T_b}$ and takes the form
\begin{align}
	\mP\Big( \sqrt{T_b} > \widetilde{C}\, N^{-1/2} + t \Big) ~\leq~ \exp(-C \,t^2 N )
\end{align}
for all $t>0$, which leads to an upper bound on the quantile\footnote{%
The concentration inequality results in a bound on the quantile $q^\infty_{1-\alpha}$ obtained with infinitely many bootstrapped statistics.
\citet[Lemma 21]{kim2023differentially} then guarantees that this translates to a bound on the quantile $q_{1-\alpha}$ obtained with a finite number of bootstrapped statistics.
For example, as illustrated in \citet[Equation 63]{kim2023differentially}, if this number is larger than $6\log(2/\beta)/\alpha$ then $q_{1-\alpha}\leq q^\infty_{1-\alpha/6}$ and the bound directly applies to the quantile with finitely many bootstrapped statistics noting that $\log(6/\alpha)\lesssim \log(1/\alpha)$ with $\log(1/\alpha)\geq 1$.
So, even though the quantile bound (which holds with probability at least $1-\beta/2$) does not at first appear to depend on $\beta$, this dependence is hidden in the condition on the number of bootstrapped statistics.
This reasoning holds throughout the proofs of this section and we do not explicitly mention the condition on the number of bootstrapped statistics (\eg, greater than $6\log(2/\beta)/\alpha$) every time.
}
\begin{align}
	q_{1-\alpha} ~\leq~ C_2 \sqrt{\frac{1}{N} \log \biggl(\frac{1}{\alpha}\biggr)},
\end{align}
holding with probability at least $1-\beta/2$.
This concentration bound can be obtained when using permutations for MMD and HSIC \citep[Lemmas 10 and 12]{kim2023differentially}, and when using wild bootstrap for MMD, HSIC and KSD (Rademacher chaos concentration of \citealp[Corollary 3.2.6]{victor1999decoupling}).

With these results, we obtain type II error control
\begin{equation}
\begin{aligned} 
	& \mP\left(\sqrt{V_0} \leq q_{1-\alpha}\right) \\
	\leq ~ &  \mP\left(\kdisc \leq q_{1-\alpha} + C_1 \sqrt{\frac{1}{N} \log \left(\frac{1}{\beta}\right)} \right) + \beta/2 \\ 
	\leq ~ &  \mP\left(\kdisc \leq C_2 \sqrt{\frac{1}{N} \log \biggl(\frac{1}{\alpha}\biggr)} + C_1 \sqrt{\frac{1}{N} \log \biggl(\frac{1}{\beta}\biggr)} \right) + \beta \\ 
	= ~ & \beta
\end{aligned}
\end{equation}
provided that 
\begin{align} 
	\kdisc ~\gtrsim~  
\sqrt{\frac{\max\!\big\{\!\log(1/\alpha), \, \log(1/\beta)\big\}}{N}},
\end{align}
as desired.

\paragraph{U-statistic.}
Consider the U-statistic $U_0\coloneqq U_k$ for some fixed kernel $k$ (\eg, \citealp[Equations 12, 26 and 52]{schrab2025practical}), which is an estimate of some kernel squared discrepancy $\kdisc^2$ (\eg, MMD$^2$, HSIC$^2$, KSD$^2$) for which $\kdisc=0$ characterises the null.
We also consider bootstrapped statistics ${U_1},\dots,{U_B}$ (either via permutations or wild bootstrap, \Cref{subsec:level_hp}), and the $(1-\alpha)$-quantile $q_{1-\alpha}$ of ${U_0},\dots,{U_B}$.
Again, the aim is to bound the probability of type II error $\PP(T_0 \leq q_{1-\alpha})$ by $\beta$ provided that the discrepancy $\kdisc$ is greater than some rate to be determined.
This can be derived by first using inequalities linking the U-statistic and V-statistic, and then using using some exponential concentration bounds for the bootstrapped U-statistic.

For completeness, even though it is not needed in this proof, we state the exponential concentration bound for the original unbiased U-statistic $U_0$  (estimating $\kdisc^2$), which takes the form
\begin{align}
\mP\Big( \big|{U_0} - \kdisc^2 \big| > t \Big) ~\leq~ \exp(-C \,t^2 N )
\end{align}
for all $t>0$, giving
\begin{align}
\big|{U_0} - \kdisc^2 \big| ~\leq~ C_1 \sqrt{\frac{1}{N} \log \biggl(\frac{1}{\beta}\biggr)}
\end{align}
with probability at least $1-\beta/2$.
For MMD, HSIC and KSD, this holds by Hoeffding's inequality \citep[Equation 5.7]{hoeffding63} provided that the kernels are bounded.

Firstly, we note that U-statistics and V-statistics are related as
\begin{align}
\label{eq:uv}
U_0 ~\geq~ V_0 - \frac{2K}{N}
\end{align}
where $K$ is a bound on the kernel.
For MMD and HSIC, this holds by \citet[Appendix E.13 and Lemma 22]{kim2023differentially}. 
For KSD, or any one-sample second-order U/V-statistic, this can be seen directly.

Secondly, the concentration for the bootstrapped statistic ${U_b}$ takes the form
\begin{align}
	\mP\Big( {U_b} > t \Big) ~\leq~ \exp(-C \,t N )
\end{align}
for all $t>0$, this gives an upper bound on the quantile
\begin{align}
	q_{1-\alpha} ~\leq~ C_3 {\frac{1}{N} \log \biggl(\frac{1}{\alpha}\biggr)},
\end{align}
holding with probability at least $1-\beta/2$.
For permutations (MMD and HSIC), this concentration bound for permuted U-statistics holds by \citep[Theorems 6.1, 6.2 \& 6.3]{kim2020minimax}.
For the wild bootstrap method (MMD, HSIC and KSD), the Rademacher chaos concentration of \citet[Corollary 3.2.6]{victor1999decoupling} gives the desired quantile bound.
See also the details of deriving the quantile bounds in \Cref{subsec:l2_proof}, 
\citet[Proposition 4]{schrab2021mmd},
\citet[Theorem 3.1]{schrab2022ksd}, and
\citet[Lemma 2]{schrab2022efficient}.

With these results, we obtain type II error control
\begin{equation}
\label{eq:urateproof}
\begin{aligned} 
	& \mP\left({U_0} \leq q_{1-\alpha}\right) \\
	\leq ~ &  \mP\left(V_0 \leq \frac{2K}{N} + q_{1-\alpha} \right) \\ 
	\leq ~ &  \mP\left(V_0 \leq \frac{2K}{N} + C_3 {\frac{1}{N} \log \biggl(\frac{1}{\alpha}\biggr)} \right) + \beta/2\\ 
	= ~ &  \mP\left(\sqrt{V_0} \leq C_4 \sqrt{\frac{1}{N} \log \biggl(\frac{1}{\alpha}\biggr)} \right) + \beta/2\\ 
	\leq ~ &  \mP\left(\kdisc \leq C_4 \sqrt{\frac{1}{N} \log \biggl(\frac{1}{\alpha}\biggr)} + C_1 \sqrt{\frac{1}{N} \log \left(\frac{1}{\beta}\right)} \right) + \beta \\ 
	= ~ & \beta
\end{aligned}
\end{equation}
provided that 
\begin{align} 
	\kdisc ~\gtrsim~  
\sqrt{\frac{\max\!\big\{\!\log(1/\alpha), \, \log(1/\beta)\big\}}{N}},
\end{align}
as desired.

\subsection{Proof sketch of differentially private kernel separation}
\label{subsec:dp_kernel_proof}

We detail the proof structure of the differentially private kernel separation results:
\Cref{eq:mmd_dp_ker} proved in \citet[Theorem 7]{kim2023differentially},
and \Cref{eq:hsic_kernel_dp} proved in \citet[Theorem 12]{kim2023differentially}.
We focus on the V-statistic case $\sqrt{V_0}$ with $V_0\coloneqq{V_k}$ for some kernel $k$, with bootstrapped statistics $\sqrt{V_1},\dots,\sqrt{V_B}$.

The proof structure is exactly the same as outlined in \Cref{subsec:kernel_proof} but taking into account the Laplace privatisation noise $\zeta_0,\dots,\zeta_B$ independently injected into $\sqrt{V_0},\dots,\sqrt{V_B}$. 
The noise is scaled by $2\Delta/\xi$ where the global sensitivity satisfies $\Delta \lesssim 1/N$ for the MMD and HSIC square-rooted V-statistics \citep[Lemmas 5 and 6]{kim2023differentially}.
The quantile $\widetilde{q}_{1-\alpha}$ of the privatised statistics is upper bounded by the sum of, the quantile of the statistics, and of the quantile of the Laplacian privatisation noise, we get
\begin{align}
	\widetilde{q}_{1-\alpha} ~\leq~ C_1 \sqrt{\frac{1}{N} \log \biggl(\frac{1}{\alpha}\biggr)} + C_2 \frac{1}{N\xi}\log \biggl(\frac{1}{\alpha}\biggr),
\end{align}
which holds with probability at least $1-\beta/3$.
The above is derived by using a closed form on the cumulative distribution function (CDF) of the Laplace distribution 
$
	F^{-1}_{\zeta}(p) \coloneqq  - \mathrm{sign}(p - 0.5)  \log(1 - 2|p - 0.5|)$ for $p \in (0,1)$.
Then, as before, we need 
\begin{align}
\big|\sqrt{V_0} - \kdisc \big| ~\leq~ C_3 \sqrt{\frac{1}{N} \log \biggl(\frac{1}{\beta}\biggr)}
\end{align}
to hold, this time with probability at least $1-\beta/3$, and also
\begin{align}
-\zeta_0 ~\leq~ C_4 \log \biggl(\frac{1}{\beta}\biggr)
\end{align}
to hold with probability at least $1-\beta/3$ (again using the closed form of the Laplace CDF).
Combining all these results we get
\begin{equation}
\begin{aligned} 
	& \mP\left(\sqrt{V_0} +\frac{2\Delta}{\xi} \zeta_0 \leq \widetilde{q}_{1-\alpha}\right) \\
	\leq ~ &  \mP\left(\kdisc+\frac{2\Delta}{\xi} \zeta_0 \leq \widetilde{q}_{1-\alpha} + C_3 \sqrt{\frac{1}{N} \log \left(\frac{1}{\beta}\right)} \right) + \frac{\beta}{3} \\ 
	\leq ~ &  \mP\left(\kdisc \leq C_1 \sqrt{\frac{1}{N} \log \biggl(\frac{1}{\alpha}\biggr)} + C_2 \frac{1}{N\xi}\log \biggl(\frac{1}{\alpha}\biggr)+ C_3 \sqrt{\frac{1}{N} \log \biggl(\frac{1}{\beta}\biggr)} -\frac{2\Delta}{\xi} \zeta_0\right) + \frac{2\beta}{3} \\ 
	\leq ~ &  \mP\left(\kdisc \leq C_1 \sqrt{\frac{1}{N} \log \biggl(\frac{1}{\alpha}\biggr)} + C_2 \frac{1}{N\xi}\log \biggl(\frac{1}{\alpha}\biggr)+ C_3 \sqrt{\frac{1}{N} \log \biggl(\frac{1}{\beta}\biggr)} + C_5 \frac{1}{N\xi} \log \biggl(\frac{1}{\beta}\biggr)\right) + \beta \\ 
	= ~ & \beta
\end{aligned}
\end{equation}
provided that 
\begin{align} 
	\kdisc ~\gtrsim~  
\max \Biggl\{ \sqrt{\frac{\max\!\big\{\!\log(1/\alpha), \, 	
		\log(1/\beta)\big\}}{N}}, \, \frac{\max\!\big\{\!\log(1/\alpha), \, \log(1/\beta)\big\}}{N\xi}  \Biggr\},
\end{align}
as desired.

\subsection{Proof sketch of robust kernel separation}
\label{subsec:robust_kernel_proof}

We detail the proof structure of the robust kernel separation results:
\Cref{eq:mmd_kernel_robust} proved in \citet[Theorem 1.i]{schrab2024robust},
and \Cref{eq:hsic_kernel_robust} proved in \citet[Theorem 2.i]{schrab2024robust}.
We focus on the V-statistic case $\sqrt{V_0}$ with $V_0\coloneqq{V_k}$ for some kernel $k$, with bootstrapped statistics $\sqrt{V_1},\dots,\sqrt{V_B}$.

The proof structure follows closely the one outlined in \Cref{subsec:kernel_proof} but, for the robust tests, the quantile is shifted by a factor of $2r\Delta$ where $r$ is the robustness parameter and $\Delta$ is the global sensitivity, which for MMD and HSIC square-rooted V-statistics, scales as $1/N$ as shown in \citet[Lemmas 5 and 6]{kim2023differentially}.
Adapting the reasoning of \Cref{subsec:kernel_proof}, we then get
\begin{equation}
\begin{aligned} 
	& \mP(\sqrt{V_0} \leq q_{1-\alpha} + 2r\Delta) \\
	\leq ~ &  \mP\left(\kdisc \leq C_1 \sqrt{\frac{1}{N} \log \biggl(\frac{1}{\alpha}\biggr)}  + C_2 \sqrt{\frac{1}{N} \log \biggl(\frac{1}{\beta}\biggr)}+ 2r\Delta\right) + \beta \\ 
	\leq ~ &  \mP\left(\kdisc \leq C_1 \sqrt{\frac{1}{N} \log \biggl(\frac{1}{\alpha}\biggr)}  + C_2 \sqrt{\frac{1}{N} \log \biggl(\frac{1}{\beta}\biggr)}+ C_3\frac{r}{N}\right) + \beta \\ 
	= ~ & \beta
\end{aligned}
\end{equation}
provided that 
\begin{align} 
	\kdisc ~\gtrsim~  
\max \biggl\{ \sqrt{\frac{\max\{\log(1/\alpha),\log(1/\beta)\}}{N}}, \ \frac{r}{N} \biggr\} 
\end{align}
as desired.

\subsection{Proof sketch of pooled kernel separation}
\label{subsec:pooled_proof}

We detail the proof structure of the pooled kernel separation results:
\Cref{eq:mmd_kernel_pooled} (fuse variant also proved in \citealp[Theorems 2 and 3]{biggs2023mmd}) and
\Cref{eq:hsic_kernel_pooled,eq:ksd_kernel_pooled,eq:std_ker_hsic_mean,eq:ksd_kernel_pooled_mean,eq:mmd_kernel_mean}.
For simplicity, consider the tests to have square-rooted V-statistic $(T_0)_k \coloneqq \sqrt{V_k}$ with kernel parameter $k\in\mathcal{K}$.
We show that the type II error is guaranteed to be controlled by $\beta$ for all alternatives satisfying
\begin{align}
\label{eq:mean_pooling_case}
\underset{k\in \Kcal}{\mathrm{mean}} ~
\kdisc_k
~\gtrsim~
\sqrt{\frac{\max\!\big\{\!\log(1/\alpha), \, \log(1/\beta), \, \log(|\Kcal|)\big\}}{N}}
\end{align}
for the pooled test over a collection $\Kcal$ with mean pooling function, and for all alternatives satisfying
\begin{align}
\max_{k\in \Kcal}\,
\kdisc_k
~\gtrsim~
\sqrt{\frac{\max\!\big\{\!\log(1/\alpha), \, \log(1/\beta), \, \log(|\Kcal|)\big\}}{N}}
\end{align}
for the pooled test over $\Kcal$ with fuse or max pooling function.
The uniform separation rate of \Cref{eq:mean_pooling_case} holds trivially from the result for fixed kernel (\Cref{subsec:kernel_proof}) and the fact the mean of discrepancies is equal to the discrepancy computed with a mean kernel \citep[Equation 78]{schrab2025practical}, which is due to the linearity of the discrepancy in the kernel.
For fuse and max kernel pooling, it is enough to prove that
type II error control by $\beta$ is guaranteed when
\begin{align}
\label{eq:reduce_to_pool}
\underset{k\in\Kcal}{\mathrm{pool}}~\kdisc_k
~\gtrsim~
\sqrt{\frac{\max\!\big\{\!\log(1/\alpha), \, \log(1/\beta), \, \log(|\Kcal|)\big\}}{N}},
\end{align}
where `$\mathrm{pool}$' is the corresponding pooling function (\ie, fuse or max).
Indeed, when using the fuse pooling function, leveraging the relation between fuse and max \citep[Equation 78]{schrab2025practical}, the `$\mathrm{pool}$' function in \Cref{eq:reduce_to_pool} can be replaced by `$\max$' resulting in an additional additive term $\log(|\Kcal|)/\nu$ term on the right hand side of \Cref{eq:reduce_to_pool}, which is absorbed in the rate provided that $\nu\geq\max(N,\log(|\Kcal|))$ (as this implies $\log(|\Kcal|)/\nu\leq \sqrt{\log(|\Kcal|)/N}$).

Then, to prove \Cref{eq:reduce_to_pool}, following the reasoning of \Cref{subsec:kernel_proof}, it suffices to show that
\begin{equation}
\big|\underset{k\in\Kcal}{\mathrm{pool}}~(T_0)_k - \underset{k\in\Kcal}{\mathrm{pool}}~\kdisc_k \big| ~\leq~ C_1 \sqrt{\frac{1}{N} \log \biggl(\frac{|\Kcal|}{\beta}\biggr)}
\qquad\textrm{ and }\qquad
	q_{1-\alpha} ~\leq~ C_2 \sqrt{\frac{1}{N} \log \biggl(\frac{|\Kcal|}{\alpha}\biggr)},
\end{equation}
each holding with probability at least $1-\beta/2$, where $q_{1-\alpha}$ is the quantile of the pooled bootstrapped statistics. 
Equivalently, as we are working with square-rooted V-statistics, we need to show that, for all $t>0$, we have
\begin{equation}
\mP\Big(\, \big|\underset{k\in\Kcal}{\mathrm{pool}}~(T_0)_k - \underset{k\in\Kcal}{\mathrm{pool}}~\kdisc_k \big| > \widetilde{t}\, \Big) ~\leq~ |\Kcal|\exp(-C \,t^2 N )
\end{equation}
and
\begin{equation}
\mP\Big(\, \underset{k\in\Kcal}{\mathrm{pool}}~(T_b)_k > \widetilde{t} \,\Big) ~\leq~ |\Kcal|\exp(-\widetilde{C} \,t^2 N ),
\end{equation}
for the cases where `$\mathrm{pool}$' is `fuse', and is `max', where $\widetilde{t} \coloneqq C' N^{-1/2} +t $ as in \Cref{subsec:kernel_proof}.

\paragraph{Maximum pooling.}
As in \Cref{subsec:kernel_proof}, each of the $|\Kcal|$ single tests satisfies
\begin{equation}
\big|(T_0)_k - \kdisc_k \big| ~\leq~ C_k \sqrt{\frac{1}{N} \log \biggl(\frac{|\Kcal|}{\beta}\biggr)},
\end{equation}
each with probability at least $1-\beta/|\Kcal|$,
for some constants $C_k$, $k\in\Kcal$, and in particular is holds true with the same constant $C=\max_{k\in\Kcal} C_k$.
Hence, with probability at least $1-\beta$, we have
\begin{equation}
\big|(T_0)_k - \kdisc_k \big| ~\leq~ C \sqrt{\frac{1}{N} \log \biggl(\frac{|\Kcal|}{\beta}\biggr)},
\end{equation}
for every $k\in\Kcal$.
Let $k_T = \mathrm{argmax}_{k\in\Kcal}\, (T_0)_k$ and $k_K = \mathrm{argmax}_{k\in\Kcal}\, \kdisc_k$
We prove that 
\begin{equation}
\label{eq:max_pool_proof}
\big|\underset{k\in\Kcal}{\mathrm{max}}~(T_0)_k - \underset{k\in\Kcal}{\mathrm{max}}~\kdisc_k \big| ~=~ \big|(T_0)_{k_T} - \kdisc_{k_K}\big| ~\leq~ C\sqrt{\frac{1}{N} \log \biggl(\frac{|\Kcal|}{\beta}\biggr)}.
\end{equation}
For simplicity we write $\delta = C\sqrt{\log (|\Kcal|/\beta)/N}$.

If $\kdisc_{k_K}> (T_0)_{k_T} + \delta$, then since $(T_0)_{k_K} + \delta \geq \kdisc_{k_K}$, we get $(T_0)_{k_K} \geq  \kdisc_{k_K} - \delta > (T_0)_{k_T}$, which contradicts the definition of $k_T = \mathrm{argmax}_{k\in\Kcal}\, (T_0)_k$. We deduce that $\kdisc_{k_K}\leq (T_0)_{k_T} + \delta$.

If $\kdisc_{k_K}< (T_0)_{k_T} - \delta$, then as $(T_0)_{k_T} - \delta \leq \kdisc_{k_T}$, we get $\kdisc_{k_K}< \kdisc_{k_T}$ which contradicts the definition of $k_K = \mathrm{argmax}_{k\in\Kcal}\, \kdisc_k$. We deduce that $\kdisc_{k_K}\geq (T_0)_{k_T} - \delta$.

This proves that \Cref{eq:max_pool_proof} holds.

For the quantile bound, we have
\begin{equation}
\label{eq:max_quantile_proof}
\mP\Big( \underset{k\in\Kcal}{\mathrm{max}}~(T_b)_k > \widetilde{t}\, \Big) 
~=~
\mP\left(\bigcup_{k\in\Kcal} \Big\{(T_b)_k > \widetilde{t}\,\Big\} \right) 
~\leq~ 
\sum_{k\in\Kcal} \mP\Big((T_b)_k > \widetilde{t}\,\Big) 
~\leq~ 
|\Kcal|\exp(-\widetilde{C} \,t^2 N )
\end{equation}
for all $t>0$,
where $\widetilde{C} = \min_{k\in\Kcal}\widetilde{C}_k$, where $\mP\big((T_b)_k > \widetilde{t}\, \big) \leq \exp(-\widetilde{C}_k \,t^2 N )$ for all $t>0$ as in \Cref{subsec:kernel_proof}, and where $\widetilde{t} \coloneqq C' N^{-1/2} +t $.

\paragraph{Fuse pooling.}
Again, as in \Cref{subsec:kernel_proof}, starting from
\begin{equation}
\big|(T_0)_k - \kdisc_k \big| ~\leq~ C \sqrt{\frac{1}{N} \log \biggl(\frac{|\Kcal|}{\beta}\biggr)},
\end{equation}
for every $k\in\Kcal$, which holds with probability at least $1-\beta$, and 
writing $\delta = C\sqrt{\log (|\Kcal|/\beta)/N}$, we then have
\begin{equation}
\label{eq:fuse_proof}
\begin{aligned}
\underset{k\in\Kcal}{\mathrm{fuse}}~(T_0)_k 
~&=~ 
\frac{1}{\nu} \log \left( \frac{1}{|\Kcal|}\sum_{k\in\Kcal} \exp \big(\nu (T_0)_k \big)  \right)\\
~&\leq~ 
\frac{1}{\nu} \log \left( \frac{1}{|\Kcal|}\sum_{k\in\Kcal} \exp \big(\nu (\kdisc_k +\delta) \big)  \right)\\
~&=~ 
\delta + \frac{1}{\nu} \log \left( \frac{1}{|\Kcal|}\sum_{k\in\Kcal} \exp \big(\nu \kdisc_k \big)  \right)\\
~&=~ 
\delta +
\underset{k\in\Kcal}{\mathrm{fuse}}~\kdisc_k 
\end{aligned}
\end{equation}
and similarly for the other direction (simply swapping the roles of $(T_0)_k$ and $\kdisc_k$), we deduce that
\begin{equation}
\label{eq:fuse_concentration_statistic}
\big|\underset{k\in\Kcal}{\mathrm{fuse}}~(T_0)_k - \underset{k\in\Kcal}{\mathrm{fuse}}~\kdisc_k \big| ~\leq~ C \sqrt{\frac{1}{N} \log \biggl(\frac{|\Kcal|}{\beta}\biggr)}.
\end{equation}
Recall from \citet[Equation 78]{schrab2025practical} that
\begin{equation}
\underset{k\in\Kcal}{\mathrm{fuse}}~(T_b)_k ~\leq~ \underset{k\in\Kcal}{\mathrm{max}}~(T_b)_k.
\end{equation}
Using this fact, for the quantile bound, we obtain
\begin{equation}
\mP\Big( \underset{k\in\Kcal}{\mathrm{fuse}}~(T_b)_k > \widetilde{t}\, \Big) 
~\leq~ 
\mP\Big( \underset{k\in\Kcal}{\mathrm{max}}~(T_b)_k > \widetilde{t}\, \Big) 
~\leq~ 
|\Kcal|\exp(-\widetilde{C} \,t^2 N )
\end{equation}
for all $t>0$, where $\widetilde{t} \coloneqq C' N^{-1/2} +t $.

\paragraph{Mean pooling.}
A similar analysis can be done for the mean pooling function. 
While this is not necessary since we can get the uniform separation rate without the additional $\log(|\Kcal|)$ term as discussed above in \Cref{eq:mean_pooling_case}, we present it nonetheless as this will be important for the global sensitivity discussion below.

As in the other cases, with probability at least $1-\beta$, we have
\begin{equation}
\big|(T_0)_k - \kdisc_k \big| ~\leq~ C \sqrt{\frac{1}{N} \log \biggl(\frac{|\Kcal|}{\beta}\biggr)},
\end{equation}
for every $k\in\Kcal$.
Writing $\delta = C\sqrt{\log (|\Kcal|/\beta)/N}$, the mean can then be bounded as
\begin{equation}
\label{eq:mean_proof}
\frac{1}{|\Kcal|}\sum_{k\in\Kcal}(T_0)_k 
~\leq ~
\frac{1}{|\Kcal|}\sum_{k\in\Kcal} \Big(\kdisc_k + \delta\Big)
~\leq~ 
\delta + \frac{1}{|\Kcal|}\sum_{k\in\Kcal} \kdisc_k.
\end{equation}
Proceeding similarly for the other direction, we conclude that 
\begin{equation}
\big|\underset{k\in\Kcal}{\mathrm{mean}}~(T_0)_k - \underset{k\in\Kcal}{\mathrm{mean}}~\kdisc_k \big| ~\leq~ C \sqrt{\frac{1}{N} \log \biggl(\frac{|\Kcal|}{\beta}\biggr)}.
\end{equation}
The quantile bound follows as the one for the maximum case in \Cref{eq:max_quantile_proof}, that is
\begin{equation}
	\mP\Big( \underset{k\in\Kcal}{\mathrm{mean}}~(T_b)_k > \widetilde{t}\, \Big) 
~\leq~ 
\mP\left(\bigcup_{k\in\Kcal} \Big\{(T_b)_k > \widetilde{t}\,\Big\} \right) 
~\leq~ 
|\Kcal|\exp(-\widetilde{C} \,t^2 N )
\end{equation}
for all $t>0$, where $\widetilde{t} \coloneqq C' N^{-1/2} +t $.

\paragraph{Global sensitivity of pooled statistic for robustness guarantees.}
Suppose that the statistic $S_k$ has global sensitivity $\Delta$ \citep[Definition 2]{kim2023differentially}, that is
\begin{align}
\bigl| S_k(\mathbb{X}_{N}^{\pi}) - S_k(\widetilde{\mathbb{X}}_{N}^{\pi}) \bigr| 
~\leq~ 
\Delta 
\end{align}
for any two datasets $\mathbb{X}_{N}$ and $\widetilde{\mathbb{X}}_{N}$ differing only in one entry, and any data permutation $\pi$.
In order to use kernel pooling under the robustness constraint, we need to study the global sensitivity of the pooled statistics for the three pooling functions (\ie, mean, max, fuse), of the form
\begin{align}
\bigl| \underset{k\in\Kcal}{\mathrm{pool}}~S_k(\mathbb{X}_{N}^{\pi}) - \underset{k\in\Kcal}{\mathrm{pool}}~S_k(\widetilde{\mathbb{X}}_{N}^{\pi}) \bigr| 
~\leq~ 
\Delta_\mathrm{pool} 
\end{align}
for any $\mathbb{X}_{N}$, $\widetilde{\mathbb{X}}_{N}$, $\pi$, as above.
Adapting the reasoning of \Cref{eq:max_pool_proof,eq:mean_proof,eq:fuse_proof}, we conclude that
$\Delta_\mathrm{mean} \leq \Delta$,
$\Delta_\mathrm{max} \leq \Delta$, and
$\Delta_\mathrm{fuse} \leq \Delta$.
In the (non-pooled) robust kernel separation setting of \Cref{subsec:robust_kernel_proof}, we observe that the quantity $r\Delta$ leads to the term $r/N$ in the kernel separation. 
In the pooled setting the quantity $\Delta_\mathrm{pool}$ leads to the same term $r/N$ for all three pooling mechanisms.
We deduce that kernel pooling for robust tests leads to the uniform separation rate
\begin{align}
\underset{k\in \Kcal}{\mathrm{mean}} ~ \mathrm{Kdisc}_k
~\gtrsim~
\max \biggl\{ \sqrt{\frac{\max\{\log(1/\alpha),\log(1/\beta),\log(|\Kcal|)\}}{N}}, \ \frac{r}{N} \biggr\}
\end{align}
for mean kernel pooling, and to the uniform separation rate
\begin{align}
\max_{k\in \Kcal}\,\mathrm{Kdisc}_k
~\gtrsim~
\max \biggl\{ \sqrt{\frac{\max\{\log(1/\alpha),\log(1/\beta),\log(|\Kcal|)\}}{N}}, \ \frac{r}{N} \biggr\}
\end{align}
for fuse/max kernel pooling.

\subsection{Proof sketch of efficient kernel separation}
\label{subsec:efficient_kernel_proof}

We detail the proof structure of the efficient kernel separation results:
\Cref{eq:mmd_kernel_efficient,eq:hsic_kernel_efficient,eq:ksd_kernel_efficient}.

We consider a block B-statistic \citep[Equation 70]{schrab2025practical} with $B$ blocks, each of size $\lfloor N/B\rfloor^2$ (we ignore the last remaining smaller block), which takes the form
\begin{equation}
U_{\textrm{block}}= \frac{1}{B}\sum_{b=1}^B U\pp{X_{1+(b-1)\lfloor N/B\rfloor},\dots, X_{b\lfloor N/B\rfloor}}
\eqqcolon \frac{1}{B}\sum_{b=1}^B U^{(b)}
\end{equation}
where $U^{(1)},\dots,U^{(B)}$ are U-statistics on $\lfloor N/B\rfloor$ samples.
The proof strategy follows the one of \Cref{subsec:kernel_proof} for the complete U-statistic, with the aim to relate $U_{\textrm{block}}$ to a non-negative $V_{\textrm{block}}=\frac{1}{B}\sum_{b=1}^B V^{(b)}$ with $V^{(b)}\coloneqq V\pp{X_{1+(b-1)\lfloor N/B\rfloor},\dots, X_{b\lfloor N/B\rfloor}}$ for $b=1,\dots,B$.
As in \Cref{eq:uv}, by \citet[Appendix E.13 and Lemma 22]{kim2023differentially} for MMD and HSIC (and by direct computation for KSD), with sample size $\lfloor N/B\rfloor$, we have
\begin{align}
U^{(b)} ~\geq~ V^{(b)} - \frac{2K}{\lfloor N/B\rfloor}
\end{align}
for $b=1,\dots,B$, where $K$ is a bound on the kernel.
We deduce that there exists some constant $C_1>0$ (depending on $K$) such that
\begin{align}
\label{eq:uv1}
U_{\textrm{block}} ~\geq~ V_{\textrm{block}} - C_1\frac{B}{N}.
\end{align}

As mentioned in \Cref{subsec:level_hp}, efficient tests using incomplete U-statistics with design $\Dcal$ rely on the wild bootstrap to avoid having to compute new entries of the kernel/core matrix.
The quantile obtained by wild bootstrap can be bounded with probability at least $1-\beta/2$ as
\begin{align}
q_{1-\alpha} 
~\lesssim~ {\frac{N}{|\Dcal|} \log \biggl(\frac{1}{\alpha}\biggr)}
\end{align}
which follows from the concentration bound for i.i.d.\,Rademacher chaos of \citet[Corollary 3.2.6]{victor1999decoupling}, see \citet[Lemma 2 and Appendix F.4]{schrab2022efficient} for details.
In this case, we work with a B-statistic with design of size $|\Dcal|=B\lfloor N/B\rfloor^2\asymp N^2/B$, so we get
\begin{align}
\label{eq:uv2}
q_{1-\alpha} 
~\leq~ C_2{\frac{B}{N} \log \biggl(\frac{1}{\alpha}\biggr)}
\end{align}
with probability at least $1-\beta/2$.

The concentration inequalities for $\sqrt{V^{(1)}},\dots,\sqrt{V^{(B)}}$ of \Cref{eq:concentration_sqrt_V} give
\begin{align}
\kdisc ~\leq~ \sqrt{V^{(b)}} + C_3 \sqrt{\frac{1}{N} \log \biggl(\frac{B}{\beta}\biggr)},
\end{align}
which hold simultaneously for $b=1,\dots,B$ with probability at least $1-\beta/2$.
Using Jensen's inequality (finite form), we deduce that, with probability at least $1-\beta/2$, it holds that
\begin{equation}
\label{eq:uv3}
\begin{aligned}
\kdisc ~&\leq~ \frac{1}{B}\sum_{b=1}^B\sqrt{V^{(b)}} + C_3 \sqrt{\frac{1}{N} \log \biggl(\frac{B}{\beta}\biggr)}\\
&\leq~ \sqrt{\frac{1}{B}\sum_{b=1}^BV^{(b)}} + C_3 \sqrt{\frac{1}{N} \log \biggl(\frac{B}{\beta}\biggr)}\\
&=~ \sqrt{V_{\textrm{block}}} + C_3 \sqrt{\frac{1}{N} \log \biggl(\frac{B}{\beta}\biggr)}\\
&\leq~ \sqrt{V_{\textrm{block}}} + C_3 \sqrt{\frac{B}{N} \log \biggl(\frac{1}{\beta}\biggr)}.
\end{aligned}
\end{equation}

Then, similarly to the complete U-statistic reasoning of \Cref{eq:urateproof}, by combining \Cref{eq:uv1,eq:uv2,eq:uv3}, we get that
\begin{equation}
\begin{aligned} 
	& \mP\left(U_{\textrm{block}} \leq q_{1-\alpha}\right) \\
	\leq ~ &  \mP\left(V_{\textrm{block}} \leq C_1\frac{B}{N} + q_{1-\alpha} \right) \\ 
	\leq ~ &  \mP\left(V_{\textrm{block}} \leq C_1\frac{B}{N} + C_2 {\frac{B}{N} \log \biggl(\frac{1}{\alpha}\biggr)} \right) + \beta/2\\ 
	= ~ &  \mP\left(\sqrt{V_{\textrm{block}}} \leq C_4 \sqrt{\frac{B}{N} \log \biggl(\frac{1}{\alpha}\biggr)} \right) + \beta/2\\ 
	\leq ~ &  \mP\left(\kdisc \leq C_4 \sqrt{\frac{B}{N} \log \biggl(\frac{1}{\alpha}\biggr)} + C_3 \sqrt{\frac{B}{N} \log \left(\frac{1}{\beta}\right)} \right) + \beta \\ 
	= ~ & \beta
\end{aligned}
\end{equation}
provided that 
\begin{align} 
\kdisc ~\gtrsim~  
\sqrt{\frac{B \max\!\big\{\!\log(1/\alpha), \, \log(1/\beta)\big\}}{N}},
\end{align}
as desired.

\subsection{Proof sketch of pooled efficient kernel separation}
\label{subsec:pooled_efficient_proof}

The pooled efficient kernel separation results of \Cref{eq:mmd_kernel_efficient_pooled,eq:hsic_kernel_efficient_pooled,eq:ksd_kernel_efficient_pooled} can simply be obtained by combining the reasoning of the pooled and efficient kernel separation results in \Cref{subsec:efficient_kernel_proof,subsec:pooled_proof}, respectively.

\subsection{Proof sketch of L2 separation}
\label{subsec:l2_proof}

We detail the proof structure of the $L^2$ separation results:
\Cref{eq:mmd_l2} proved in \citet[Corollary 7]{schrab2021mmd}, 
\Cref{eq:hsic_l2} proved in \citet[Theorem 3]{schrab2022efficient}
(extension of \citealp[Corollary 2,]{albert2019adaptive} with theoretical quantiles, and of \citealp[Proposition 8.7,]{kim2020minimax} with permuted quantiles),
and \Cref{eq:ksd_pow_l2} proved here.
Here, we improve the dependence in $\beta$ to be logarithmic for all these results.
For L2 separation, we focus on the U-statistic case with $U_0\coloneqq U_k$ for some kernel $k$, with bootstrapped statistics ${U_1},\dots,{U_B}$.

\subsubsection{MMD two-sample testing}
\label{subsubsec:mmd_l2_proof}
We prove the MMD case in details, framing the reasoning in a general setting that will easily be adapted to the HSIC and KSD cases.
Recall that we assume that the kernel $k$ integrates to 1 and takes the form $\kk(x,y) \coloneqq \prod_{i=1}^d K_i\big((x_i-y_i)/{\lambda_i}\big)\big/{\lambda_i}$ for $x,y\in\R^d$ and $\lambda_i>0$, $i=1,\dots,d$.
The kernel integral transform $S_\lambda$ is defined as
$
\p{S_\lambda f}(y) 
\coloneqq \int_{\R^d} \kk(x,y) f(x) \,\dd x
$,
$y\in\R^d$ for any function $f:\R^d\to\R$.
Recall that $h_\lambda(x, x', y, y') \coloneqq k(x,x') + k(y,y') - k(x,y') - k(x',y)$ for any $x,y,x',y'\in\R^d$.
We denote the difference in densities by $\psi=p-q$.
We focus on the U-statistic case with $U_0\coloneqq U_{\kl}$ as defined in \citet[Equation 12]{schrab2025practical}.

\paragraph{Statistic concentration.}
Using Bernstein inequality for U-statistic of \citet{arcones1995bernstein} as presented by \citet[Theorem 2]{peel2010empirical} (as opposed to Chebyshev's inequality in \citealp[Appendix E.2]{schrab2021mmd}), we obtain
\begin{equation}
\label{eq:bernstein}
\big|U_0 - \kdisc^2\big| 
~\lesssim~
\sqrt{\frac{\sigma_1^2}{N}\log\left(\frac{1}{\beta}\right)} + \frac{1}{N}\log\left(\frac{1}{\beta}\right)
\end{equation}
with probability at least $1-\beta/2$,
where
\begin{equation}
\label{eq:sigma1}
\sigma_{1}^2 
~\coloneqq~ 
\VV{Z}{\EE{Z'}{\h(Z,Z')}}\\
~\lesssim~ 
\norm{S_\lambda \psi}_{L^2}^2
\end{equation}
as shown in \citet[Appendix E.3]{schrab2021mmd}.

\paragraph{Quantile bound.}
For either permutations or wild bootstrap, we can also obtain a quantile bound \citep[Appendix E.4]{schrab2021mmd} of the form 
\begin{equation}
\label{eq:1q}
\PPP\p{\big|U_b\big|~\geq~ \frac{\tilde{C}_1}{N} \sqrt{\frac{1}{N(N-1)}\sum_{1\leq i\neq j \leq N} \h(Z_i, Z_j)^2}\log\p{\frac{1}{\alpha}} \,\bigg|\, \Xn,\Yn}~\leq~ \alpha.
\end{equation}
Using Bernstein inequality for U-statistic \citep[Theorem 2]{peel2010empirical} (as opposed to Markov's inequality in \citealp[Appendix E.4]{schrab2021mmd}), we get
\begin{align}
\frac{1}{N(N-1)}\sum_{1\leq i\neq j \leq N} \h(Z_i, Z_j)^2
~&\leq~
\mathbb{E}\left[\frac{1}{N(N-1)}\sum_{1\leq i\neq j \leq N} \h(Z_i, Z_j)^2\right] + \sqrt{\frac{\widetilde\sigma_1^2}{N}\log\left(\frac{1}{\beta}\right)} + \frac{1}{N}\log\left(\frac{1}{\beta}\right)
\nonumber\\
~&\lesssim~
\mathbb{E}\left[\kk(Z, Z')^2\right] +\sqrt{\frac{1}{N(\LL)^2}\log\left(\frac{1}{\beta}\right)} + \frac{1}{N}\log\left(\frac{1}{\beta}\right)\nonumber\\
~&\lesssim~
\frac{1}{\LL}+\frac{1}{\LL}\sqrt{\frac{1}{N}\log\left(\frac{1}{\beta}\right)}+ \frac{1}{N}\log\left(\frac{1}{\beta}\right)\nonumber\\
~&\lesssim~
\frac{1}{\LL}\log\left(\frac{1}{\beta}\right)\label{eq:quant_bound_last_line}
\end{align}
with probability at least $1-\beta/2$, where
\begin{equation}
\label{eq:3q}
\begin{aligned}
\widetilde{\sigma}_{1}^2 
~&\coloneqq~ 
\VV{Z}{\EE{Z'}{\h(Z,Z')^2}}
~\lesssim~ 
\EE{Z}{\p{\EE{Z'}{\h(Z,Z')^2}}^2}
~\lesssim~ 
\EE{Z}{\p{\EE{Z'}{\kk(Z,Z')^2}}^2}\\
~&\lesssim~ 
\EE{Z}{\p{\frac{1}{\LL}}^2}
~=~ 
\frac{1}{(\LL)^2}.
\end{aligned}
\end{equation}
This means that
\begin{equation}
\label{eq:4q}
\PPP\p{\abs{U_b}\geq \frac{\widetilde{C}_2}{N\sqrt{\LL}} \sqrt{\log\left(\frac{1}{\beta}\right)} \log\p{\frac{1}{\alpha}} \,\bigg|\, \Xn,\Yn}\leq \alpha,
\end{equation}
or equivalently
\begin{align} 
\label{eq:quantile_berstein}
{q}_{1-\alpha}~ &\lesssim~ \frac{1}{N\sqrt{\LL}} \sqrt{\log\left(\frac{1}{\beta}\right)} \log\p{\frac{1}{\alpha}},
\end{align} 
with probability at least $1-\beta/2$.

\paragraph{Kdisc separation.}
Now, the type II error can be controlled as
\begin{equation}
\label{eq:trick_cancel}
\begin{aligned} 
	& \mP(U_0 \leq q_{1-\alpha}) \\
	\leq ~ &  \mP\left(\kdisc^2 ~\leq~ C_1 \sqrt{\frac{\norm{S_\lambda \psi}_{L^2}^2}{N}\log\left(\frac{1}{\beta}\right)} + \frac{C_2}{N}\log\left(\frac{1}{\beta}\right) + q_{1-\alpha}\right) + \beta/2 \\ 
	\leq ~ &  \mP\left(\kdisc^2 ~\leq~ \frac{1}{2}\norm{S_\lambda \psi}_{L^2}^2 + \widetilde{C}_1 {\frac{1}{N}\log\left(\frac{1}{\beta}\right)} + \frac{C_2}{N}\log\left(\frac{1}{\beta}\right) + q_{1-\alpha}\right) + \beta \\ 
	\leq ~ &  \mP\left(\kdisc^2 ~\leq~ \frac{1}{2}\norm{S_\lambda \psi}_{L^2}^2 + \frac{C}{2N\sqrt{\LL}} {\log\left(\frac{1}{\beta}\right)} \log\p{\frac{1}{\alpha}}\right) + \beta \\ 
	= ~ & \beta
\end{aligned}
\end{equation}
provided that 
\begin{equation} 
\label{eq:rate_new_kdisc_l2}
	\kdisc^2 ~\geq~  
\frac{1}{2}\norm{S_\lambda \psi}_{L^2}^2 + \frac{C}{2N\sqrt{\LL}} {\log\left(\frac{1}{\beta}\right)} \log\p{\frac{1}{\alpha}}
\end{equation}
for some constant $C>0$,
where we have used the common bound $2\sqrt{xy}\leq x+y$.

\paragraph{Kdisc/L2 expression.}
The kernel discrepancy can be expressed in terms of the $L^2$ norms \citep[Appendix E.5]{schrab2021mmd}
\begin{equation}
\label{eq:mmdl2expression}
\kdisc^2 = \frac{1}{2}\left(
\big\|\psi\big\|_{L^2}^2 + \big\|S_\lambda \psi\big\|_{L^2}^2 - \big\|\psi - S_\lambda \psi\big\|_{L^2}^2
\right).
\end{equation}

\paragraph{L2 separation.}
The uniform separation of \Cref{eq:rate_new_kdisc_l2} can then be expressed as
\begin{equation} 
\label{eq:usr2_new}
\big\|\psi\big\|_{L^2}^2 ~\geq~  
\big\|\psi - S_\lambda \psi\big\|_{L^2}^2 + \frac{C}{N\sqrt{\LL}} {\log\left(\frac{1}{\beta}\right)} \log\p{\frac{1}{\alpha}}.
\end{equation}

\paragraph{Sobolev control.}
Following \citet[Appendix E.6]{schrab2021mmd}, assuming that $\psi$ lies in a Sobolev ball of smoothness $s$, the term $\big\|\psi - S_\lambda \psi\big\|_{L^2}^2$ can then be bounded as
\begin{equation}
\big\|\psi - S_\lambda \psi\big\|_{L^2}^2
~\leq~
c\big\|\psi\big\|_{L^2}^2
+
\widetilde{C} \sum_{i=1}^d \lambda^{2s}_i
\end{equation}
for some constants $c\in(0,1)$ and $\widetilde{C}>1$.
Substituting this bound in the uniform separation condition of \Cref{eq:usr2_new}, the type II error is guaranteed to be controlled by $\beta$ under the stronger requirement
\begin{equation} 
\label{eq:usr3_new}
\big\|\psi\big\|_{L^2}^2 
~\gtrsim~  
\sum_{i=1}^d \lambda^{2s}_i
+ \frac{1}{N\sqrt{\LL}} {\log\left(\frac{1}{\beta}\right)} \log\p{\frac{1}{\alpha}}.
\end{equation}

\paragraph{Optimal bandwidth.}
Finally, setting $\lambda_1=\dots=\lambda_d=\lambda$ and equating $\lambda^{2s}$ to $\lambda^{-d/2}N^{-1}\log(1/\alpha)\log(1/\beta)$ (similarly to as in \citealp[Appendix E.7]{schrab2021mmd}), we obtain $\lambda = \big(\log(1/\alpha)\log(1/\beta)/N\big)^{2/(4s+d)}$ which gives the final uniform separation over the Sobolev ball of any smoothness $s>0$
\begin{equation} 
\label{eq:usr5_new}
\big\|\psi\big\|_{L^2} 
~\gtrsim~  
\p{\frac{\log(1/\alpha)\log(1/\beta)}{N}}^{2s/(4s+d)}
\end{equation}
where $\psi=p-q$.

\subsubsection{HSIC independence testing}
We show how the HSIC result can be obtained based on the MMD reasoning detailed above in Appendix \ref{subsubsec:mmd_l2_proof}.
Recall that we assume that the kernels $\kxl(x,\widetilde{x}) \coloneqq \prod_{i=1}^{d_\mathcal{X}} K_i^\mathcal{X}\big((x_i-\widetilde{x}_i)/{\lambda_i}\big)\big/{\lambda_i}$ and $\kyl(y,\widetilde{y}) \coloneqq \prod_{j=1}^{d_\mathcal{Y}} K_j^\mathcal{Y}\big((y_j-\widetilde{y}_j)/{\mu_j}\big)\big/{\mu_j}$ with bandwidths $\lambda\in (0,\infty)^{d_{\mathcal{X}}}$ and $\mu\in (0,\infty)^{d_{\mathcal{Y}}}$, both integrate to 1.
The kernel integral transform $S_{\lambda,\mu}$ is defined as
$
\p{S_{\lambda,\mu} f}(x,y) 
\coloneqq \int_{\R^{d_\mathcal{X}}}\int_{\R^{d_\mathcal{Y}}} f(\widetilde{x},\widetilde{y}) \kxl(x,\widetilde{x}) \kyl(y,\widetilde{y}) \,\dd \widetilde{y} \,\dd \widetilde{x}
$,
$(x,y)\in\R^{d_\mathcal{X}}\times\R^{d_\mathcal{Y}}$ for any function $f:\R^{d_\mathcal{X}}\times\R^{d_\mathcal{Y}}\to\R$.
Recall that $h_{\lambda,\mu}$ is defined as in \citet[Equation 23]{schrab2025practical}.
We denote the difference between the joint and the product of marginals by $\psi=p_{xy}-p_x\otimes p_y$.
We focus on the U-statistic case with $U_0\coloneqq U_{\kxl,\kyl}$ as defined in \citet[Equation 26]{schrab2025practical}.

The exact same reasoning as the one presented for MMD holds for HSIC using $\lambda_1\cdots\lambda_{d_\mathcal{X}}\mu\cdots\mu_{d_\mathcal{Y}}$ instead of $\lambda_1\cdots\lambda_d$ (for details, see \citealp[Theorem 3]{schrab2022efficient} which extends the results of \citealp[Corollary 2,]{albert2019adaptive} and \citealp[Section 8.5]{kim2020minimax}), with the only difference being in the derivation of the quantile bound, which we present in details here.
The aim is to prove that, with probability at least $1-\beta/2$, we have
\begin{align} 
	{q}_{1-\alpha}~ &\lesssim~ \frac{1}{N\sqrt{\lambda_1\cdots\lambda_{d_\mathcal{X}}\mu\cdots\mu_{d_\mathcal{Y}}}} {\log\left(\frac{1}{\beta}\right)} \log\p{\frac{1}{\alpha}}.
\end{align} 
For this we derive the quantile bound as in \citet[Theorem 3]{schrab2022efficient} but relying on Bernstein's inequality to obtain the desired logarithmic depdendence on $\beta$.

As in \citet[Appendix F.5.1]{schrab2022efficient}, applying the exponential concentration bound of \citet[Theorem 6.3]{kim2020minimax}, which is based on \citet[Theorem 4.1.12]{pena1999decoupling}, we obtain
\begin{equation}
q_{1-\alpha} ~\lesssim~ \max\p{
    \frac{\Sigma}{N}\ln\pp{\frac{1}{\alpha}},
    \frac{M}{N^{3/2}}\ln\pp{\frac{1}{\alpha}}^{3/2}
},
\end{equation}
where
\begin{align}
    M ~\coloneqq~ \max_{1\leq i,j,r,s \leq N} \big|k_\lambda(X_i,X_j)\ell_\mu(Y_r,Y_s)\big| 
    ~\lesssim~ \frac{1}{\lambda_1\cdots\lambda_{d_\mathcal{X}}\mu\cdots\mu_{d_\mathcal{Y}}}
\end{align}
and
\begin{align}
&\Sigma^2 
~\coloneqq~ 
\p{\frac{1}{N^2}
\sum_{1\leq i,j\leq N}
k_\lambda(X_i,X_j)^2
}
\p{\frac{1}{N^2}
\sum_{1\leq i,j\leq N}
\ell_\mu(Y_i,Y_j)^2
}
.
\end{align}
Using the same reasoning as in the MMD case relying on Bernstein's inequality, we obtain that
\begin{align}
\frac{1}{N^2}
\sum_{1\leq i,j\leq N}
k_\lambda(X_i,X_j)^2
&=
\frac{1}{N\lambda_1\cdots\lambda_{d_\mathcal{X}}}
+
\frac{(N-1)}{N}
\p{
\frac{1}{N(N-1)}
\sum_{1\leq i\neq j\leq N}
k_\lambda(X_i,X_j)^2
}
\lesssim
\frac{1}{\lambda_1\cdots\lambda_{d_\mathcal{X}}}\log\left(\frac{1}{\beta}\right)
\end{align}
with probability at least $1-\beta/4$, and
\begin{align}
\frac{1}{N^2}
\sum_{1\leq i,j\leq N}
\ell_\mu(X_i,X_j)^2
~&=~
\frac{1}{N\mu\cdots\mu_{d_\mathcal{Y}}}
+
\frac{(N-1)}{N}
\p{
\frac{1}{N(N-1)}
\sum_{1\leq i\neq j\leq N}
\ell_\mu(X_i,X_j)^2
}
~\lesssim~
\frac{1}{\mu\cdots\mu_{d_\mathcal{Y}}}\log\left(\frac{1}{\beta}\right)
\end{align}
with probability at least $1-\beta/4$.
We deduce that, with probability at least $1-\beta/2$, it holds
\begin{equation}
	\Sigma ~\lesssim~ \frac{1}{\sqrt{\lambda_1\cdots\lambda_{d_\mathcal{X}}\mu\cdots\mu_{d_\mathcal{Y}}}}\log\left(\frac{1}{\beta}\right).
\end{equation}
with probability at least $1-\beta/2$.
We conclude that
\begin{equation}
\label{eq:quant_bound_hsic_}
q_{1-\alpha} 
~\lesssim~ 
\frac{\log(1/\alpha)\log(1/\beta)}{N\sqrt{\lambda_1\cdots\lambda_{d_\mathcal{X}}\mu\cdots\mu_{d_\mathcal{Y}}}}
\,
\max\left\{
1,\,
\sqrt{\frac{\log(1/\alpha)}{N\lambda_1\cdots\lambda_{d_\mathcal{X}}\mu\cdots\mu_{d_\mathcal{Y}}}}
\right\}
\end{equation}
Then, following the proof structure of the MMD case from Appendix \ref{subsubsec:mmd_l2_proof}, we finally set $\lambda_1,\dots,\lambda_{d_\mathcal{X}},\mu_1\dots,\mu_{d_\mathcal{Y}}$ to all be equal to $\big(\log(1/\alpha)\log(1/\beta)/N\big)^{2/(4s+d_{\mathcal{X}}+d_{\mathcal{Y}})}$.
Then, assuming that $4s \geq d_{\mathcal{X}}+d_{\mathcal{Y}}$, we have
\begin{equation}
\lambda_1\cdots\lambda_{d_\mathcal{X}}\mu\cdots\mu_{d_\mathcal{Y}}
~=~ \p{\frac{\log(1/\alpha)\log(1/\beta)}{N}}^{2(d_{\mathcal{X}}+d_{\mathcal{Y}})/(4s+d_{\mathcal{X}}+d_{\mathcal{Y}})}
~\geq~
\frac{\log(1/\alpha)}{N},
\end{equation}
and so the quantile bound of \Cref{eq:quant_bound_hsic_} becomes
\begin{equation}
q_{1-\alpha} 
~\lesssim~ 
\frac{\log(1/\alpha)\log(1/\beta)}{N\sqrt{\lambda_1\cdots\lambda_{d_\mathcal{X}}\mu\cdots\mu_{d_\mathcal{Y}}}}.
\end{equation}
We conclude that the uniform separation over any Sobolev ball of smoothness $s\geq (d_{\mathcal{X}}+d_{\mathcal{Y}})/4$ is
\begin{equation} 
\big\|\psi\big\|_{L^2} 
~\gtrsim~  
\p{\frac{\log(1/\alpha)\log(1/\beta)}{N}}^{2s/(4s+d)}
\end{equation}
where $\psi=p_{xy}-p_x\otimes p_y$.

\subsubsection{KSD goodness-of-testing testing}
\label{subsec:ksd_l2_proof_}

\paragraph{Score.} 
For simplicity, consider the case of a model $P$ with bounded score function $\sbm_P(x) = \nabla \log(p(x))$, nonetheless, we stress that our results can hold for more general settings.
As an example, this includes the case of the multivariate t-distribution with density $p(x) \propto \p{1 + \frac{\|x\|_2^2}{\nu}}^{-\frac{\nu+d}{2}}$ for $x\in\R^d$ and $\nu>0$ degrees of freedom. It indeed has bounded score as
\begin{equation}
\sbm_P(x) 
~=~ -(\nu+d)\frac{x}{\nu + \|x\|_2^2}
\qquad\textrm{ giving }\qquad
\big\|\sbm_P(x)\big\|_2^2 
~=~ (\nu+d)\frac{\|x\|^2_2}{\big(\nu + \|x\|_2^2\big)^2}
~\leq~ \frac{\nu+d}{4\nu}
\end{equation}
for all $x\in\R^d$.
Here, we work with full support $\R^d$ for the model.
We recall that for KSD testing, we always need the support to be connected in order to avoid the blindness issue of score-based methods to mixing proportions on isolated components \citep{wenliang2020blindness,zhang2022towards}.

\paragraph{Kernel.}
Moreover, as commonly used for the KSD \citep{gorham2017measuring}, we consider the IMQ kernel
\begin{equation}
\kl(x,y) \coloneqq \frac{1}{\lambda^d} \frac{1}{\left(1+{\norm{x-y}_2^2}\big/{\lambda^2}\right)^\beta}
\end{equation}
for $x,y\in\R^d$, with bandwidth $\lambda>0$ and $\beta\in(\nicefrac{1}{2},1)$.
We consider $\beta$ to be fixed and to not track its dependence in the bounds.
The choice $\beta\in(\nicefrac{1}{2},1)$ ensures that the kernel takes the form $\kl(x,y) = \frac{1}{\lambda^d} K\!\left(\frac{x-y}{\lambda}\right)$, where $K$ integrates to some constant value (\ie, it can be normalised to integrate to 1 if desired). 
For all $x,y\in\R^d$, this kernel satisfies
\begin{align}
\left|\frac{\partial}{\partial x_i}\kl(x,y)\right|
~&=~
\left|
\frac{2\beta(x_i-y_i)}{\lambda^2+\|x-y\|_2^2}
\right|
\kl(x,y)
~\lesssim~
\frac{1}{\lambda}\,\kl(x,y),
\\
\left|\frac{\partial}{\partial y_i}\kl(x,y)\right|
~&\lesssim~
\frac{1}{\lambda}\,\kl(x,y),
\\
\left|\frac{\partial}{\partial x_i\partial y_i}\kl(x,y)\right|
~&\lesssim~
\frac{1}{\lambda^2}\,\kl(x,y).
\end{align}
From these, we deduce that
\begin{align}
\big\|\nabla_{\!1} \kl(x,y)\big\|_2^2 
~&\lesssim~
\frac{1}{\lambda^{2d}}\,\kl(x,y)^2,\\
\big\|\nabla_{\!2} \kl(x,y)\big\|_2^2 
~&\lesssim~
\frac{1}{\lambda^{2d}}\,\kl(x,y)^2,\\
\big\|\nabla_{\!1}^\top \!\big(\nabla_{\!2}\kl(x,y)\big)\big\|_2^2 
~&\lesssim~ 
\frac{1}{\lambda^{4d}}\kl(x,y)^2,
\end{align}
where we recall that $\nabla_{\!1}^\top \!\big(\nabla_{\!2}k(x,y)\big) = \sum_{i=1}^d \frac{\partial}{\partial y_i}\frac{\partial}{\partial x_i}k(x,y)$.
Recall from \citet[Section 2.3]{schrab2025practical} that the Stein kernel takes the form
\begin{equation}
h_P(x,y) \coloneqq
k(x,y)\sbmp(x)^\top\sbmp(y)
+ \big(\nabla_{\!1}k(x,y)\big)^\top  \sbmp(Y)
+ \big(\nabla_{\!2}k(x,y)\big)^\top  \sbmp(X)
+ \nabla_{\!1}^\top \big(\nabla_{\!2}k(x,y)\big).
\end{equation}
Using Cauchy--Schwartz inequality, together with the above bounds on the derivatives of the kernel, as well as on the score function, we obtain that
\begin{equation}
\label{eq:h_k_link}
h_\lambda(x,y)^2 ~\lesssim~ \frac{1}{\lambda^{4d}}\kl(x,y)^2
\end{equation}
for all $x,y\in\R^d$.
\Cref{eq:h_k_link} pinpoints the main difference between the MMD and KSD cases, the proof structure will be the same but the additional scaling in $\lambda$ will be affect the rate and needs to be kept track of. 

\paragraph{Kdisc/L2 expression.}
Using the result of \citet[Theorem 3.6]{liu2016kernelized}, we have
\begin{equation}
\label{eq:ksdl2}
\begin{aligned}
\mathrm{KSD}^2 
&= \E_{Q,Q}\!\left[
\big(\sbm_P(X)-\sbm_Q(X)\big)^\top \big(\sbm_P(Y)-\sbm_Q(Y)\big)\,
\kk(X,Y)
\right] \\
&= \int_{\R^d} \int_{\R^d} \Big(q(x)\big(\sbm_P(x)-\sbm_Q(x)\big)\Big)^\top \Big(q(y)\big(\sbm_P(y)-\sbm_Q(y)\big)\Big) \,\kk(x,y)\, \dd x \dd y\\
&= \int_{\R^d} \psib(x)^\top \!\left(\int_{\R^d}  \psib(y) \,\kk(x,y) \,\dd y \right) \dd x \\
&= \int_{\R^d} \psib(x)^\top \!\left(S_{\lambda}\psib\right)\!(x) \,\dd x \\
&= \Big\langle \psib, S_\lambda\psib\Big\rangle_{L^2} \\
&= \frac{1}{2} \p{\norm{\psib}_{L^2}^2 + \norm{S_{\lambda}\psib}_{L^2}^2 - \norm{\psib - S_{\lambda}\psib}_{L^2}^2},
\end{aligned}
\end{equation}
similar to the MMD expression of \Cref{eq:mmdl2expression} but with a different $\psib$ function.
Since the KSD is a score-based discrepancy, we will characterise departures from the null with the difference in scores multiplied by the data density
\begin{equation}
\psib(x) ~\coloneqq~ \Big(\nabla\log p(x) - \nabla\log q(x)\Big) q(x).
\end{equation}
We guarantee high test power for the KSD test against all alternatives with $\big\|\psib\big\|_{L^2}$ greater than some rate to be determined, with regularity condition that $\psib = (\nabla\log p - \nabla\log q) \,q$ belongs to a Sobolev ball.
Note that the separation is quantified in the metric
\begin{equation}
\big\|\psib\big\|_{L^2}^2 = \int_{\R^d} \norm{\sbm_P(x)-\sbm_Q(x)}^2_2 \,q(x)^2 \, \dd x,
\end{equation}
also considered by \citet[Proposition 3.3]{liu2016kernelized},
which is closely related to the Fisher divergence \citep{johnson2004information}
\begin{equation}
\int_{\R^d} \norm{\sbm_P(x)-\sbm_Q(x)}^2_2 \,q(x) \, \dd x.
\end{equation}

\paragraph{Statistic concentration.}
The statistic concentration is the same as \Cref{eq:bernstein} for the MMD case, that is
\begin{equation}
\big|U_0 - \kdisc^2\big| 
~\lesssim~
\sqrt{\frac{\sigma_1^2}{N}\log\left(\frac{1}{\beta}\right)} + \frac{1}{N}\log\left(\frac{1}{\beta}\right)
\end{equation}
with probability at least $1-\beta/2$.
However, the bound on $\sigma_{1}^2\coloneqq\VV{Z}{\EE{Z'}{\h(Z,Z')}}$ differs.
Ideally, we would like to upper bound $\sigma_{1}^2$ by  
$
\norm{S_\lambda \psib}_{L^2}^2 = 
\int_{\R^d}\big\|
\EE{Y}{ k(x,Y)\dbm(Y)}
\big\|^2_2 \,\dd x
$, similarly to \Cref{eq:sigma1} for the MMD case.
However, we can only show 
$
\sigma_{1}^2 
\lesssim 
\int_{\R^d}\EE{Y}{\|k(x,Y) \dbm(Y)\|_2^2}
 \,\dd x
$ which is not tight enough.
Hence, we simply upper bound $\sigma_{1}^2$ in terms of $\lambda$, using \Cref{eq:h_k_link} we get
\begin{equation}
\sigma_{1}^2 
~\lesssim~
\EE{Z}{\p{\EE{Z'}{\h(Z,Z')}}^2}
~\lesssim~
\EE{Z,Z'}{\h(Z,Z')^2}
~\lesssim~
\frac{1}{\lambda^{4d}}
\EE{Z,Z'}{\kl(Z,Z')^2}
~\lesssim~
\frac{1}{\lambda^{5d}}.
\end{equation}
We deduce that
\begin{equation}
\label{eq:quant_ksd_bound}
\big|U_0 - \kdisc^2\big| 
~\lesssim~
\frac{1}{\lambda^{5d/2}}\,\sqrt{\frac{1}{N}\log\left(\frac{1}{\beta}\right)} + \frac{1}{N}\log\left(\frac{1}{\beta}\right)
\end{equation}
with probability at least $1-\beta/2$.
We note that without the bound in 
$
\norm{S_\lambda \psib}_{L^2}^2 
$, we cannot use the trick of \Cref{eq:trick_cancel} to cancel the 
$
\norm{S_\lambda \psib}_{L^2} 
$
terms and get a rate in $N^{-1}$ instead of $N^{-1/2}$.

\paragraph{Quantile bound.}
Adapting the reasoning of 
\Cref{eq:1q,eq:quant_bound_last_line,eq:3q,eq:4q,eq:quantile_berstein} to keep track of the new bandwidth scaling of \Cref{eq:h_k_link}, we get that \Cref{eq:quant_bound_last_line} becomes
\begin{align}
\frac{1}{N(N-1)}\sum_{1\leq i\neq j \leq N} \h(Z_i, Z_j)^2
~&\leq~
\mathbb{E}\left[\h(Z, Z')^2\right] + \sqrt{\frac{\widetilde\sigma_1^2}{N}\log\left(\frac{1}{\beta}\right)} + \frac{1}{N}\log\left(\frac{1}{\beta}\right)
\nonumber\\
~&\lesssim~
\frac{1}{\lambda^{4d}}\,\mathbb{E}\left[\kk(Z, Z')^2\right] +\sqrt{\frac{1}{N\lambda^{10d}}\log\left(\frac{1}{\beta}\right)} + \frac{1}{N}\log\left(\frac{1}{\beta}\right)\nonumber\\
~&\lesssim~
\frac{1}{\lambda^{5d}}+\frac{1}{\lambda^{5d}}\sqrt{\frac{1}{N}\log\left(\frac{1}{\beta}\right)}+ \frac{1}{N}\log\left(\frac{1}{\beta}\right)\nonumber\\
~&\lesssim~
\frac{1}{\lambda^{5d}}\log\left(\frac{1}{\beta}\right)\label{eq:quant_bound_last_line_ksd}
\end{align}
with probability at least $1-\beta/2$, with
\begin{equation}
\begin{aligned}
\widetilde{\sigma}_{1}^2 
~&\coloneqq~ 
\VV{Z}{\EE{Z'}{\h(Z,Z')^2}}
~\lesssim~ 
\EE{Z}{\p{\EE{Z'}{\h(Z,Z')^2}}^2}
~\lesssim~ 
\frac{1}{\lambda^{8d}}\,\EE{Z}{\p{\EE{Z'}{\kk(Z,Z')^2}}^2}
~\lesssim~ 
\frac{1}{\lambda^{10d}}.
\end{aligned}
\end{equation}
Finally, following the reasoning of 
\Cref{eq:1q,eq:quant_bound_last_line,eq:3q,eq:4q,eq:quantile_berstein}, we obtain the quantile bound
\begin{align} 
\label{eq:real_quanti}
{q}_{1-\alpha}~ &\lesssim~ \frac{1}{N\lambda^{5d/2}} \sqrt{\log\left(\frac{1}{\beta}\right)} \log\p{\frac{1}{\alpha}}
\end{align} 
holding with probability at least $1-\beta/2$.

\paragraph{Kdisc separation.}
Using \Cref{eq:quant_ksd_bound,eq:real_quanti}, the type II error can be controlled as
\begin{equation}
\begin{aligned} 
	& \mP(U_0 \leq q_{1-\alpha}) \\
	\leq ~ &  \mP\left(\kdisc^2 ~\leq~  \frac{C_1}{\lambda^{5d/2}} \sqrt{\frac{1}{N}\log\left(\frac{1}{\beta}\right)} + \frac{C_2}{N}\log\left(\frac{1}{\beta}\right) + q_{1-\alpha}\right) + \beta/2 \\ 
	\leq ~ &  \mP\left(\kdisc^2 ~\leq~  \frac{C_1}{\lambda^{5d/2}} \sqrt{\frac{1}{N}\log\left(\frac{1}{\beta}\right)} + \frac{C_2}{N}\log\left(\frac{1}{\beta}\right) + \frac{C_3}{N\lambda^{5d/2}} \sqrt{\log\left(\frac{1}{\beta}\right)} \log\p{\frac{1}{\alpha}}
\right) + \beta/2 \\ 
		\leq ~ &  \mP\left(\kdisc^2 ~\leq~ \frac{C_4}{\lambda^{5d/2}\sqrt{N}} {\log\left(\frac{1}{\beta}\right)} \log\p{\frac{1}{\alpha}}\right) + \beta \\ 
	= ~ & \beta
\end{aligned}
\end{equation}
provided that 
\begin{equation} 
\label{eq:cond_power_ksd_kdisc}
\kdisc^2 ~\gtrsim~  
\frac{1}{\lambda^{5d/2}\sqrt{N}} {\log\left(\frac{1}{\beta}\right)} \log\p{\frac{1}{\alpha}}.
\end{equation}

\paragraph{L2 separation.}
Using the expression of \Cref{eq:ksdl2}, the power guaranteeing condition of \Cref{eq:cond_power_ksd_kdisc} becomes
\begin{equation} 
\label{eq:usr2_new_ksd}
\big\|\psib\big\|_{L^2}^2 ~\gtrsim~  
\big\|\psib - S_\lambda \psib\big\|_{L^2}^2 
-\norm{S_{\lambda}\psib}_{L^2}^2 
+
\frac{C}{\lambda^{5d/2}\sqrt{N}} {\log\left(\frac{1}{\beta}\right)} \log\p{\frac{1}{\alpha}}.
\end{equation}
The difference with the MMD case (\emph{e.g.}, \Cref{eq:usr2_new}) is the rate $N^{-1/2}$ instead of $N^{-1}$, the extra term $-\norm{S_{\lambda}\psib}_{L^2}^2$, and the rate in $\lambda$.
Here, we simply bound $-\norm{S_{\lambda}\psib}_{L^2}^2$ by 0 (instead of being able to use this term to improve the rate in $N$ as in the MMD case), we get
\begin{equation} 
\label{eq:usr2_new_ksd2}
\big\|\psib\big\|_{L^2}^2 ~\gtrsim~  
\big\|\psib - S_\lambda \psib\big\|_{L^2}^2 
+
\frac{C}{\lambda^{5d/2}\sqrt{N}} {\log\left(\frac{1}{\beta}\right)} \log\p{\frac{1}{\alpha}}.
\end{equation}

\paragraph{Sobolev control.}
We assume that $\psib$ belongs to a Sobolev ball of smoothness $s$, that is
\begin{equation}
\int_{\R^d} \big\|\xi\big\|^{2s}_2 \,\big\|\widehat \psib(\xi)\big\|^2_2 \,\dd \xi ~\leq~ (2\pi)^d
\end{equation}
where $\widehat \psib$ is a vector of Fourier transforms of the form $\widehat \psib(\xi) \coloneqq \displaystyle\int_{\R^d} \psib(x) e^{-ix\!^\top\!\xi} \,\dd x$ for all $\xi\in\R^d$. 
Then, following the reasoning of \citet[Appendix E.6]{schrab2021mmd},\footnote{The proof is presented for a kernel $\prod_{i=1}^d K_i\!\left(\frac{x_i-y_i}{\lambda_i}\right)\!\big/\lambda_i$, which is a product of one-dimensional translation invariant kernels, in order to allow for different bandwidths in each dimension.
When using the same bandwidth in all dimensions, the proof can easily be adapted to hold for any translation invariant kernel $K\!\left(\frac{x-y}{\lambda}\right)\!\big/\lambda^d$ using the same reasoning.} we obtain that,
assuming $\psib$ lies in a Sobolev ball of smoothness $s$,
we get the same bound
\begin{equation}
\big\|\psib - S_\lambda \psib\big\|_{L^2}^2
~\leq~
c\big\|\psib\big\|_{L^2}^2
+
\widetilde{C} \sum_{i=1}^d \lambda^{2s}_i.
\end{equation}
Then, the overall uniform separation rate of
\Cref{eq:usr2_new_ksd2}
becomes
\begin{equation} 
\label{eq:usr3_new_new}
\big\|\psib\big\|_{L^2}^2 
~\gtrsim~  
\sum_{i=1}^d \lambda^{2s}_i
+
\frac{C}{\lambda^{5d/2}\sqrt{N}} {\log\left(\frac{1}{\beta}\right)} \log\p{\frac{1}{\alpha}}.
\end{equation}

\paragraph{Optimal bandwidth.}
Equating the terms $\lambda^{2s}$ and $\lambda^{-5d/2}N^{-1/2}\log(1/\alpha)\log(1/\beta)$, 
we obtain the bandwidth $\lambda = \big(\log(1/\alpha)\log(1/\beta)/\sqrt{N}\big)^{2/(4s+5d)}$, giving the uniform separation over the Sobolev ball of smoothness $s$,
\begin{equation} 
\big\|\psib\big\|_{L^2} 
~\gtrsim~  
\p{\frac{\log(1/\alpha)\log(1/\beta)}{\sqrt N}}^{2s/(4s+5d)}
\end{equation}
characterised with respect to
$
\psib = (\nabla\log p - \nabla\log q) \,q
$.

\subsection{Proof sketch of efficient L2 separation}
\label{subsec:eff_l2_proof}

We detail the proof structure of the efficient $L^2$ separation results:
\Cref{eq:efficient_l2,eq:hsic_l2_eff} proved in \citet[Theorem 1]{schrab2022efficient}, 
and \Cref{eq:ksd_l2_eff} proved here.

To unify the HSIC case with the MMD and KSD cases, we let $d=d_{\mathcal{X}}+d_{\mathcal{Y}}$ and $\lambda_{i+d_{\mathcal{X}}} \coloneqq \mu_i$ for $i=1,\dots,d_{\mathcal{Y}}$ so that $\lambda_1\cdots\lambda_{d_\mathcal{X}}\mu\cdots\mu_{d_\mathcal{Y}}=\LL$, this way all three cases can be treated with the same notation.
We let $U_0$ represent the MMD/HSIC/KSD incomplete U-statistic for some kernel $k$ \citep[Section 3]{schrab2025practical}.

As in the case of kernel separation (\Cref{subsec:kernel_proof}), the proof of $L^2$ separation in \Cref{subsec:l2_proof} relies on two exponential concentration results: one for the test statistic (\Cref{eq:bernstein}), and one for the bootstrapped statistic leading to a quantile bound (\Cref{eq:quantile_berstein}).
To prove the desired efficient $L^2$ separation rates of \Cref{eq:efficient_l2,eq:hsic_l2_eff,eq:ksd_l2_eff}, it then suffices to derive versions of the results of \Cref{eq:bernstein,eq:quantile_berstein} with $N$ replaced by $|\Dcal|/N$.
We now illustrate how this can be done.

An equivalent version to the exponential concentration result of \Cref{eq:bernstein} for incomplete U-statistics is provided by \citet[Theorem 3.3]{maurer2022exponential} guaranteeing that, with probability at least $1-\beta/2$, we have\footnote{%
	Referring to the notation of \citet[Theorem 3.3]{maurer2022exponential}, we have $\alpha,\beta,\gamma$ bounded by constants, $A\leq N/|\Dcal|$, $B\leq N^2/|\Dcal|^2$, $C\leq N/|\Dcal|$, where the bound for $A$ holds assuming that the number of entries of $\D$ in each row of the $N\times N$ kernel/core matrix is at most of the order of $\sqrt{|\Dcal|}$. Intuitively this requires that the entries of $\Dcal$ are spread out around the kernel matrix, they cannot all be concentrated on a same row when $|\Dcal|$ is small compared to $N^2$.
} 
\begin{equation}
\label{eq:incomplete_bernstein}
\big|U_0 - \kdisc^2 \big| 
~\lesssim~ \sqrt{\frac{N}{|\Dcal|}\sigma_1^2 \log \biggl(\frac{1}{\beta}\biggr)} + {\frac{N}{|\Dcal|} \log \biggl(\frac{1}{\beta}\biggr)} 
\end{equation}
where $\sigma_{1}^2\coloneqq\VV{Z}{\EE{Z'}{\h(Z,Z')}}$ as in \Cref{eq:sigma1}.

To derive an equivalent version of \Cref{eq:quantile_berstein}, we note that the exponential concentration bound for i.i.d. Rademacher chaos of \citet[Corollary 3.2.6]{victor1999decoupling} can be applied to the case of incomplete U-statistics (see \citealp[Theorem 1]{schrab2022efficient}) to obtain that, with probability at least $1-\beta/2$, we have
\begin{equation}
q_{1-\alpha} 
~\lesssim~ 
\sqrt{\frac{1}{|\Dcal|^2}\sum_{1\leq i\neq j\leq N} h_\lambda(Z_i,Z_j)^2}\log\p{\frac{1}{\alpha}}
~\lesssim~ 
\frac{N}{|\Dcal|}
\frac{1}{\sqrt{\LL}}
\sqrt{\log\p{\frac{1}{\beta}}}
\log\p{\frac{1}{\alpha}}
\end{equation}
using the bound of \Cref{eq:quant_bound_last_line} relying on Bernstein's inequality, where $q_{1-\alpha}$ is the $(1-\alpha)$-quantile of the incomplete bootstrapped U-statistics.

With these two exponential concentration bounds adapted for incomplete U-statistics, the efficient $L^2$ separation results of \Cref{eq:efficient_l2,eq:hsic_l2_eff,eq:ksd_l2_eff} can be proved by following the exact same resoning presented in \Cref{subsec:l2_proof}.

\subsection{Proof sketch of aggregated L2 separation}
\label{subsec:agg_l2_proof}

We detail the proof structure of the aggregated $L^2$ separation results:
\Cref{eq:mmd_l2_agg} proved in \citet[Corollary 10]{schrab2021mmd}, 
\Cref{eq:hsic_l2_agg} proved in \citet[Theorem 3]{schrab2022efficient} with estimated quantiles (and in \citealp[Corollary 3]{albert2019adaptive} with theoretical quantiles),
and \Cref{eq:ksd_l2_agg} proved here.

To unify the HSIC case with the MMD and KSD cases, we let $d=d_{\mathcal{X}}+d_{\mathcal{Y}}$ and $\lambda_{i+d_{\mathcal{X}}} \coloneqq \mu_i$ for $i=1,\dots,d_{\mathcal{Y}}$ so that $\lambda_1\cdots\lambda_{d_\mathcal{X}}\mu\cdots\mu_{d_\mathcal{Y}}=\LL$, this way all three cases can be treated with the same notation using U-statistics.
The KSD rate is slightly different, but the reasoning is the same.

Since multiple testing rejects the null if any of the adjusted tests rejects, it means that the separation rate of the aggregated test is tighter than each of the separation rates of the adjusted tests \citep[Appendix E.9]{schrab2021mmd}.
Hence, to bound the separation rate of the $\alpha$-level multiple test over a collection of bandwidths $\Lambda$, it suffices to bound the separation rate of one of the single tests for a specific bandwidth $\widetilde{\lambda}$ with adjusted level $\alpha w_{\widetilde{\lambda}}$.

Let $\lambda_1 = \dots = \lambda_d = \lambda$, and consider the aggregated test (\Cref{sec:multiple_testing}) over the bandwidths
\begin{equation}
\Lambda ~\coloneqq~ \left\{2^{-\ell}: \ell \in \left\{1,\dots, \left\lceil\frac{2}{d}\log_2\!\left(\frac{N/\log(\log(N))}{\log(1/\alpha)\log(1/\beta)}\right)\right\rceil\right\}\right\},
\end{equation}
with each bandwidth $2^{-\ell}$ having weight $w_\ell \coloneqq {6}/{\ell^{2}\pi^{2}}$, all summing to a quantity less than 1.
Consider the specific bandwidth $\widetilde\lambda = 2^{-\widetilde{\ell}}$ with
\begin{equation}
\widetilde\ell
~\coloneqq~
\left\lceil\frac{2}{4s+d}\log_2\!\left(\frac{N/\log(\log(N))}{\log(1/\alpha)\log(1/\beta)}\right)\right\rceil
~\leq~
\left\lceil\frac{2}{d}\log_2\!\left(\frac{N/\log(\log(N))}{\log(1/\alpha)\log(1/\beta)}\right)\right\rceil
\end{equation}
which satisfies
\begin{equation}
\frac{1}{2}\p{\frac{\log(1/\alpha)\log(1/\beta)}{N/\log(\log(N))}}^{2/(4s+d)}
~\leq~
\widetilde\lambda
~\leq~
\p{\frac{\log(1/\alpha)\log(1/\beta)}{N/\log(\log(N))}}^{2/(4s+d)}.
\end{equation}
This means that this specific bandwidth in the collection scales as
\begin{equation}
\label{eq:optimal_bandwidth8}
\widetilde\lambda
~\asymp~
\p{\frac{\log(1/\alpha)\log(1/\beta)}{N/\log(\log(N))}}^{2/(4s+d)}
\end{equation}
like the optimal bandwidth in \Cref{subsec:l2_proof}.
Then, as in \Cref{eq:usr3_new}, the uniform separation rate of this specific test, with bandwith $\widetilde \lambda = 2^{-\widetilde \ell}$ and adjusted level $\alpha w_{\widetilde \ell}$, is
\begin{equation} 
\big\|\psi\big\|_{L^2}^2 
~\gtrsim~  
\sum_{i=1}^d \widetilde{\lambda}^{2s}
+ \frac{1}{N{\widetilde{\lambda}^{d/2}}} {\log\left(\frac{1}{\beta}\right)} \log\p{\frac{1}{\alpha w_{\widetilde \ell}}}
\end{equation}
which holds when
\begin{equation} 
\big\|\psi\big\|_{L^2}^2 
~\gtrsim~  
\sum_{i=1}^d \widetilde{\lambda}^{2s}
+ \frac{\log(\log(N))}{N{\widetilde{\lambda}^{d/2}}} {\log\left(\frac{1}{\beta}\right)} \log\p{\frac{1}{\alpha}}
\end{equation}
since $w_{\widetilde \ell} = {6}/{{\widetilde{\ell}}^{2}\pi^{2}}$ giving
$
\ln\p{{1}/{w_{\widetilde\ell}}\,}
\lesssim
\ln(\widetilde\ell\,)
\lesssim
\ln(\ln(N)).
$
Substituting the expression of \Cref{eq:optimal_bandwidth8} for the specific bandwidth $\widetilde\lambda$ of the collection $\Lambda$, we get that the overall aggregated test over $\Lambda$ controls the type II error by $\beta$ whenever
\begin{equation} 
\big\|\psi\big\|_{L^2} 
~\gtrsim~  
\p{\frac{\log(1/\alpha)\log(1/\beta)}{N/\log(\log(N))}}^{2s/(4s+d)}
\end{equation}
for any Sobolev smoothness $s>0$ for MMD and KSD, and for any Sobolev smoothness $s\geq (d_{\mathcal{X}}+d_{\mathcal{Y}})/4$ for HSIC (as in \Cref{subsec:l2_proof}).

\subsection{Proof sketch of aggregated efficient L2 separation}
\label{subsec:agg_eff_l2_proof}

By combining the reasoning of the aggregated and efficient $L^2$ separation results in \Cref{subsec:agg_l2_proof,subsec:eff_l2_proof}, respectively, one obtains the aggregated efficient $L^2$ separation results:
\Cref{eq:mmd_l2_eff_agg} proved in \citet[Theorem 2]{schrab2022efficient}, 
\Cref{eq:hsic_l2_eff_agg} proved in \citet[Theorem 2]{schrab2022efficient},
and \Cref{eq:ksd_l2_eff_agg} proved here.

\subsection{Proof sketch of differentially private L2 separation}
\label{subsec:dp_l2_proof}

We detail the proof structure of the differentially private $L^2$ separation results:
\Cref{eq:mmd_l2_dp_low,eq:mmd_l2_dp_mid,eq:mmd_l2_dp_high} proved in \citet[Theorem 9]{kim2023differentially},
and \Cref{eq:hsic_l2_dp_low,eq:hsic_l2_dp_mid,eq:hsic_l2_dp_high} proved in \citet[Theorem 14]{kim2023differentially}.

The proof structure is similar to the one of the non-private case depicted in \Cref{subsec:l2_proof} with two major differences.
The first one is that the results need to be adapted from holding for U-statistics to holding for V-statistics, this is done by expressing the V-statistics in terms of the U-statistics \citep[Equation 38 and Lemma 22]{kim2023differentially}.
The second difference is that the added privatisation Laplacian noise needs to be taken into account, it is scaled by the global sensivity of the square-rooted V-statistic which is of order $1/(N\!\sqrt{\LL})$.
For detailed proofs, see \citet[Theorems 9 and 14]{kim2023differentially}, in these the dependence on $\beta$ is polynomial, we believe it can be improved to be logarithmic by relying on Berstein's inequality as done in \Cref{subsec:l2_proof}.

\section*{Acknowledgements}
I, Antonin Schrab, acknowledge support from the U.K.\ Research and Innovation under grant number EP/S021566/1.

\clearpage
\fancyhf{}
\fancyfoot[RO]{\thepage}
\lhead{}
\chead{A Practical Introduction to Kernel Discrepancies: MMD, HSIC \& KSD}
\cfoot{\thepage}
\rfoot{}
\bibliography{biblio}

\end{document}